\documentclass{arcpaper}

\usepackage{amsmath,amsfonts,bm}

\def\eqref#1{equation~\ref{#1}}

\def\1{\bm{1}}

\DeclareMathAlphabet{\mathsfit}{\encodingdefault}{\sfdefault}{m}{sl}
\SetMathAlphabet{\mathsfit}{bold}{\encodingdefault}{\sfdefault}{bx}{n}

\usepackage{hyperref}
\usepackage{url}
\usepackage{graphicx}
\usepackage{xspace}
\usepackage{etoolbox}
\usepackage{xcolor}
\usepackage{float}
\usepackage{wrapfig}
\usepackage{tabularx}
\usepackage[most]{tcolorbox}
\usepackage{titletoc}
\usepackage{xcolor}
\usepackage{tcolorbox}
\usepackage{xcolor}

\newcommand{\suplogo}[1]{\raisebox{0.45em}{\includegraphics[height=0.65em]{figures/#1.png}}\kern0.05em}

\definecolor{arcLink}{HTML}{4A7500}

\hypersetup{
  colorlinks=true,
  allcolors=arcLink
}
\tcbuselibrary{breakable,skins,listings}
\makeatletter
\@ifpackageloaded{FiraMono}{}{%
  \let\ARCPromptSavedTTDefault\ttdefault
  \usepackage[type1,lining]{FiraMono}
  \let\ttdefault\ARCPromptSavedTTDefault
}
\makeatother

\definecolor{ARCPromptAccent}{HTML}{76B900}
\definecolor{ARCPromptDeep}{HTML}{2F4D00}
\definecolor{ARCPromptTint}{HTML}{F1F8E6}
\definecolor{ARCPromptRule}{HTML}{A8D55E}
\newcommand{\ARCPromptFont}{\fontencoding{T1}\fontfamily{FiraMono-TLF}\fontseries{m}\fontshape{n}\fontsize{9}{11}\selectfont}
\tcbset{arc prompt/.style={%
  enhanced jigsaw,breakable,sharp corners,
  colback=white,colframe=ARCPromptRule,boxrule=0.4pt,
  colbacktitle=ARCPromptTint,coltitle=ARCPromptDeep,
  fonttitle=\ARCPromptFont\bfseries,
  left=8pt,right=8pt,top=6pt,bottom=6pt,boxsep=0pt,
  toptitle=4pt,bottomtitle=4pt,
  before skip=8pt,after skip=8pt,
  pad at break*=1.5mm}}
\newtcolorbox{ARCProsePrompt}[1]{arc prompt,
  title={#1},title after break={#1 (continued)},
  fontupper=\ARCPromptFont\fontsize{9}{10.5}\selectfont,
  before upper={\raggedright\setlength{\parindent}{0pt}\setlength{\parskip}{0pt}}}
\newtcblisting{ARCCodePrompt}[1]{arc prompt,
  title={#1},title after break={#1 (continued)},listing only,
  listing options={basicstyle=\ARCPromptFont,
    columns=fullflexible,keepspaces=true,showstringspaces=false,
    breaklines=true,breakatwhitespace=true,breakautoindent=false,
    breakindent=1em,aboveskip=0pt,belowskip=0pt,
    numbers=none,tabsize=2}}
\usepackage{booktabs}
\usepackage{colortbl}
\usepackage{xcolor,longtable,array,colortbl}
\let\OriginalTTDefault\ttdefault
\usepackage[type1,lining]{FiraMono}
\let\ttdefault\OriginalTTDefault
\definecolor{glauxgreenbg}{HTML}{F1F8E6}

\newtcolorbox{greenbox}{
  enhanced,
  breakable,
  colback=glauxgreenbg,
  colframe=glauxgreenbg,
  boxrule=0pt,
  arc=1.2mm,
  outer arc=1.2mm,
  boxsep=0pt,
  left=1.5mm,
  right=1.5mm,
  top=1.0mm,
  bottom=1.0mm,
  before skip=0pt,
  after skip=0pt,
}

\usepackage{xcolor}
\usepackage{soul}

\definecolor{tbmBackground}{HTML}{F1F8E6}

\DeclareRobustCommand{\tbm}[1]{%
  \leavevmode%
  \begingroup%
    \sethlcolor{tbmBackground}%
    \hl{#1}%
  \endgroup%
}

\newcommand{\GlauxLogo}{%
  \raisebox{-0.1em}{%
    \includegraphics[height=1.0em]{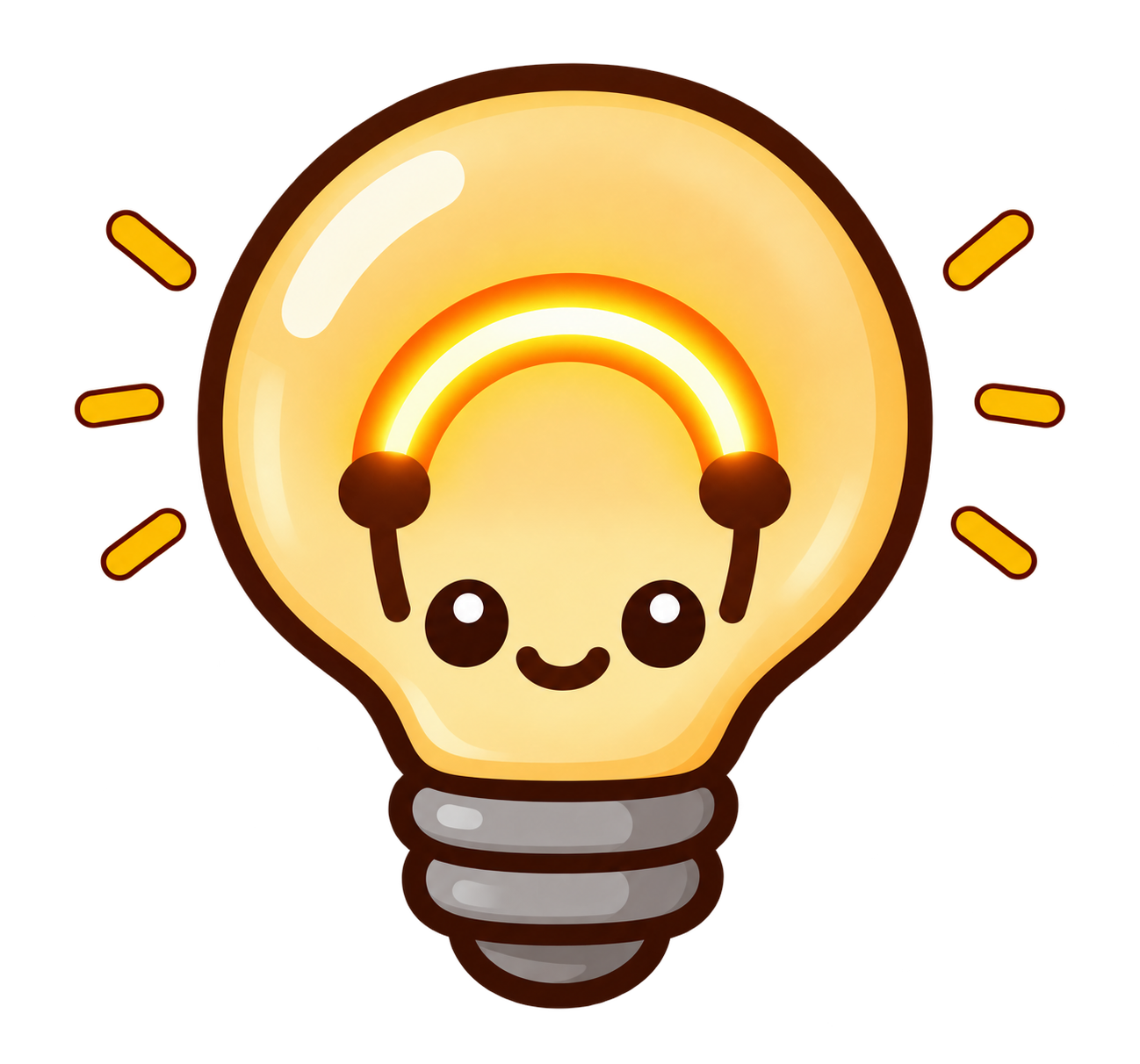}%
  }%
}

\newcommand{\name}{%
\textsc{Arc}\xspace
}

\newcommand{\tracename}{%
\textsc{Arc}-Trace\xspace
}

\newcommand{\datasetname}{%
\textsc{Arc}-Trace-DROID\xspace
}

\definecolor{uiucorange}{HTML}{FF5F05}   %
\definecolor{stanfordred}{HTML}{8C1515}  %
\definecolor{nvidiagreen}{HTML}{76B900}  %
\definecolor{usydochre}{HTML}{E64626}    %

\newcommand{\nametwo}{\textsc{Arc}}

\apptocmd{\maketitle}{}{}{}

\title{
\GlauxLogo \name: A Reasoning Recipe for Robot \\Foundation Models
}

\author[1,3]{Gokul Puthumanaillam}
\author[2,3]{Tao Sun}
\author[3]{Elie Aljalbout}
\author[3]{Moritz Reuss}
\author[3]{Zhaoshuo Li}
\author[3,4]{Fabio Ramos}
\author[3,5,\dagger]{Ankit Goyal}
\author[3,\dagger]{Jenai Xuning Yang}

\affiliation[1]{University of Illinois Urbana-Champaign}
\affiliation[2]{Stanford University}
\affiliation[3]{NVIDIA}
\affiliation[4]{University of Sydney}
\affiliation[5]{Proception}

\contribution[\dagger]{Equal advising, listed alphabetically}

\usepackage{needspace}
\BeforeBeginEnvironment{wrapfigure}{\Needspace*{0.3\textheight}}

\begin{document}

\begin{abstract}
The prevailing approach to improving robot foundation models (RFMs) relies on larger models, more robot demonstrations, and costly training at scale. We show that there exists an effective and efficient complementary approach: the right reasoning recipe can substantially improve the zero-shot task performance of existing state-of-the-art RFMs. We refer to this recipe as \GlauxLogo \textbf{\name}. It consists of three key ingredients: a reasoning trace, a scalable automatic labeling pipeline, and a strategy for adapting pretrained RFMs to use these traces for control. First, we find that effective reasoning traces should be grounded in the robot's next action and explain its causal structure: why the action is appropriate and what effect it should produce. Second, we show that these traces can be generated automatically from existing demonstrations, enabling us to construct \nametwo-Trace-DROID from DROID without collecting new robot data. Third, we show how state-of-the-art VLAs such as $\pi_{0.5}$ and WAMs such as Cosmos3-Nano-Policy can learn to use these traces for control, with fine-tuning and inference tailored to each model's architecture and capabilities.
Using \name, we obtain gains in zero-shot RFM performance that, to our knowledge, are unprecedented without additional robot demonstrations or foundation-scale training. The adapted models establish a new state of the art on RoboLab-120 and MolmoSpaces, with gains of up to 50 percentage points on RoboLab-Reasoning-50. On real robots, \name improves $\pi_{0.5}$'s task success by 82.2 percentage points.\\
\textbf{Project website:} \href{https://arc-robot-reasoning.github.io/}{https://arc-robot-reasoning.github.io/}
\end{abstract}

\maketitle

\newskip\arcsavedtextfloatsep
\arcsavedtextfloatsep=\textfloatsep
\setlength{\textfloatsep}{12pt plus 2pt minus 2pt}
\AddToHookNext{shipout/after}{\global\textfloatsep=\arcsavedtextfloatsep}
\begin{figure}[b!]
    \centering
    \captionsetup{skip=6pt}
    \includegraphics[width=0.86\textwidth]{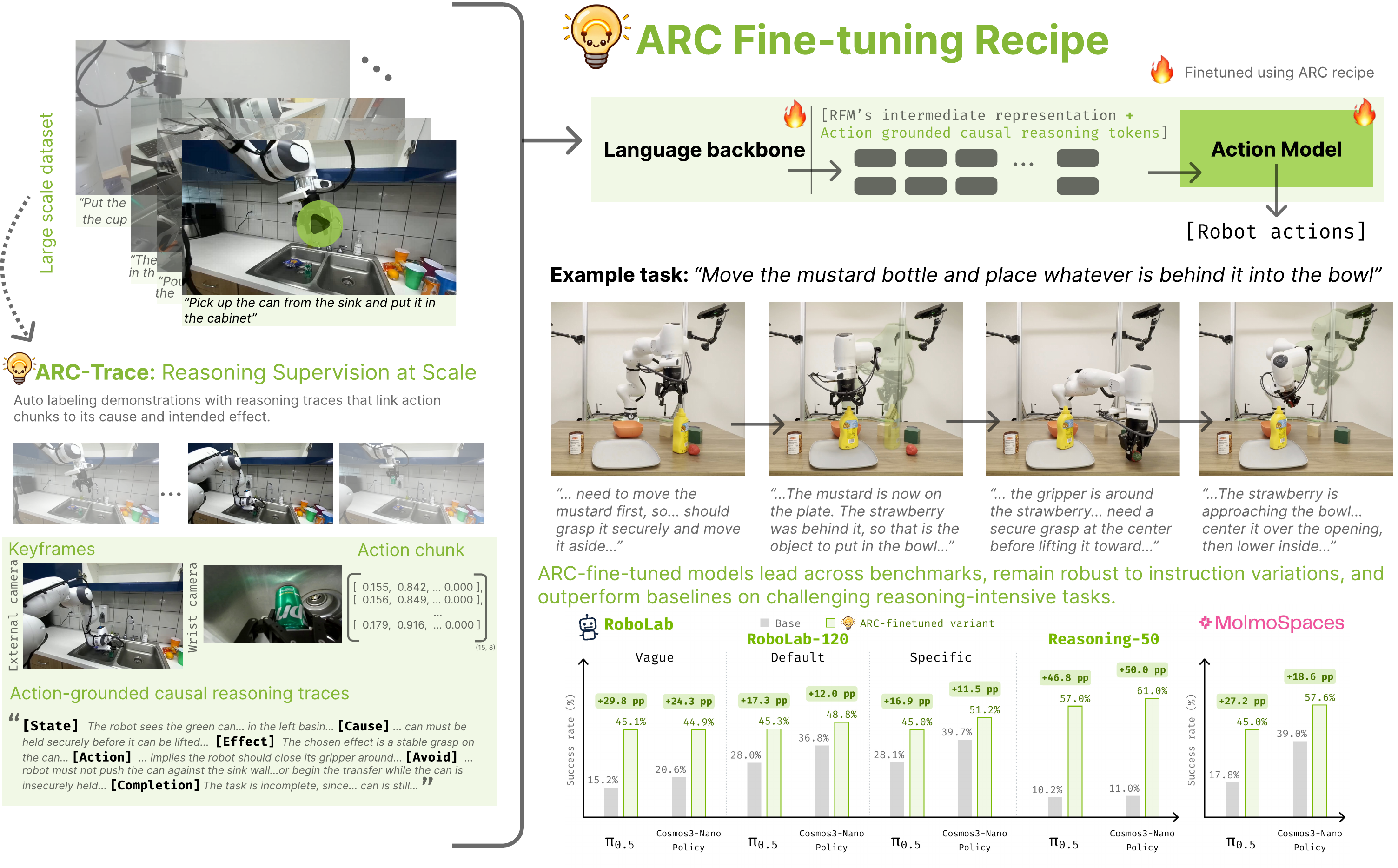}
\caption{Overview of \GlauxLogo \name: \name-Trace automatically relabels DROID with action-grounded causal traces to produce \datasetname. We show how to fine-tune VLAs and WAMs with this supervision, achieving benchmark-leading performance without additional robot demonstrations.}
\label{fig:teaser}
\end{figure}

\section{Introduction}

Robot foundation models (RFMs), trained by imitation on large demonstration datasets, have become broadly capable at manipulation \citep{openvla,cosmos3,pi0, pi05}. Yet the dominant recipe for making them better remains brute-force scaling, with more robot data, larger models, and another round of foundation-scale training. This makes every gain in task success expensive to obtain. How can we improve the performance of existing RFMs without modifying their architectures, adding new robot data, or training at the scale of foundation models?

In this work, we study this problem. Specifically, we find that adding or refining the reasoning capabilities of RFMs is an effective and efficient strategy for improving them. With the right recipe, reasoning can substantially improve the performance of existing state-of-the-art models such as $\pi_{0.5}$~\citep{pi05} and Cosmos3-Nano-Policy~\citep{cosmos3} (Fig.~\ref{fig:teaser}). To our knowledge, this is the first work to achieve such substantial gains in zero-shot RFM performance through any strategy that does not require additional robot demonstrations or foundation-scale training. We call this reasoning recipe \GlauxLogo \name.

In \name, we bring together three pieces: the right reasoning content, a scalable source of supervision, and a training procedure that adapts the model to use that reasoning for control.
First, what should the reasoning contain? We find that reasoning for control should be causal, explaining why an action is appropriate in the current situation and what effect it should produce, as well as grounded in action, specifying the concrete step the robot should take to bring about that effect. We introduce \tbm{action-grounded causal traces} that connect what is true now to what the robot should do next and what should become true as a result.
Second, how do we obtain this supervision at scale? Robot demonstrations record observations and actions but typically omit the reasoning behind individual decisions. We present an approach for automatically labeling demonstrations with causal traces aligned to action chunks. We use it to construct \tbm{\nametwo-Trace-DROID} by relabeling DROID \citep{droid}.
And third, how do we adapt models to use this reasoning for control? We show that an RFM can be fine-tuned to condition its behavior on reasoning traces while retaining its existing control capabilities. We use the same reasoning traces and training data across models and show how to tailor the fine-tuning and inference procedures to each model's architecture and capabilities.

We evaluate this recipe on two representative state-of-the-art RFMs: Cosmos3-Nano-Policy~\citep{cosmos3}, a world action model (WAM), and $\pi_{0.5}$ \citep{pi05}, a vision-language-action model (VLA). We find that \name improves both RFMs across all three settings on RoboLab-120, with the adapted models ranking first and second in each setting. On MolmoSpaces, task success improves by 27.2 and 18.6 percentage points (p.p.) for $\pi_{0.5}$ and Cosmos3-Nano-Policy. These results establish a new state of the art on RoboLab-120~\citep{robolab} and MolmoSpaces~\citep{kim2026molmospaces}. On the reasoning-focused RoboLab-Reasoning-50, \name improves task success by 46.8 p.p. for $\pi_{0.5}$ and 50.0 p.p. for Cosmos3-Nano-Policy. Further, these improvements carry over to real world execution, where the \name gains 82.2 p.p. on reasoning and long-horizon tasks. Across these evaluations, \name-trained models more reliably follow underspecified instructions, infer missing steps, complete long-horizon tasks, adapt to changes, and recover from mistakes.

We further quantify how different components of the recipe contribute to performance. Our results show that adding intermediate language without the right structure provides limited improvement; larger gains arise from combining causal reasoning content with a training procedure that grounds it in action. As a side effect, the recipe also improves training and inference efficiency. Cosmos3-Nano-Policy reaches baseline success with $4.3\times$ fewer training updates, while $\pi_{0.5}$ and Cosmos3-Nano-Policy match their base policies at inference with $10\times$ and $2\times$ fewer denoising steps.

\vspace{0\baselineskip}
\begin{greenbox}
\textbf{Contributions: }We introduce \GlauxLogo \name, a recipe for instilling reasoning into pretrained RFMs. We make four contributions.
(i) We identify {action-grounded causal traces} as an intermediate representation for robot reasoning.
(ii) We introduce ARC-Trace, a labeling pipeline that generates action-grounded causal traces aligned to action chunks and use it to create \datasetname by relabeling DROID.
(iii) We develop a strategy for adapting pretrained RFMs to use reasoning for control and show how to tailor the fine-tuning and inference procedures to each model's architecture.
(iv) We validate \name on Cosmos3-Nano-Policy and $\pi_{0.5}$ in simulation and on real-robot benchmarks, demonstrating substantial gains in zero-shot task performance that, to our knowledge, are unprecedented without additional robot demonstrations or foundation-scale training.
\end{greenbox}

\section{Related Work}
\label{sec:related}

\tbm{\textbf{Robot foundation models.}}
General-purpose robot policies have been developed by learning across tasks and embodiments \citep{gato,jang_bc-z:_2022,jiang_vima:_2023,rt1,bousmalis_robocat:_2023,oneill_open_2024} and transferring pretrained visual and linguistic knowledge to action prediction \citep{driess_palm-e:_2023,rt2,openvla,pi0,pi05}. Video-pretrained and world-action models additionally learn how the environment evolves during execution \citep{wu_unleashing_2023,cheang_gr-2:_2024,cen_worldvla:_2025,cosmos3,ye_world_2026}. Our work builds on these pretrained controllers and investigates how reasoning supervision can improve their performance using robot demonstrations already available.

\tbm{\textbf{Reasoning in foundation models.}}
Reasoning allows pretrained models to work through intermediate decisions before producing a final answer \citep{wei_chain--thought_2023,zhang_multimodal_2024}. Prior work improves this process by considering alternative solutions \citep{wang_self-consistency_2023,yao_tree_2023}, incorporating tools and feedback \citep{yao_react:_2023}, and learning from intermediate supervision \citep{zelikman_quiet-star:_2024,deepseek-ai_deepseek-r1:_2025}. In robotics, these ideas have been used in two particularly relevant ways: to determine the steps a robot should execute, and to train a policy to reason about the actions it predicts. The first includes selecting available skills to accomplish an instruction \citep{ahn_as_2022,huang_grounded_2023}, generating programs that combine perception and control \citep{liang_code_2023,singh_progprompt:_2022}, and revising plans using observations and execution feedback \citep{huang_inner_2022,zeng_socratic_2022}. The second uses reasoning annotations to teach a policy which information matters for action prediction \citep{ecot,sun2026revisiting}. Both approaches allow robots to benefit from knowledge beyond their demonstrated motions, but require the controller to understand the guidance it receives. Recent work addresses this requirement through richer steering commands and visually grounded hierarchical control \citep{chen2026steerable}. 

\tbm{\textbf{Reasoning modalities.}}
There are several ways to express the intermediate decisions needed for robotic control. Spatial representations specify where an interaction should occur through image features, action maps, or voxel representations \citep{zeng2020january,shridhar2021cliport,shridhar2023perceiver}, while keypoints and geometric constraints describe relationships that the robot should establish between objects \citep{liu2024moka,huang2023voxposer,huang2024rekep}. Motion-based representations provide guidance through trajectory sketches and visual traces \citep{gu2023rt,zheng2025tracevla,molmoact,zhong2026acot}. Predictive representations specify how the scene should evolve, using future predictions \citep{du2023learning,ko2024learning,zhou2024robodreamer, zhen20243d}. These approaches give the controller an explicit spatial or temporal target. Their connection to control depends on the representation. Geometric constraints can be solved through motion optimization \citep{huang2023voxposer,huang2024rekep}, depth and trajectory predictions can be encoded as intermediate tokens \citep{molmoact}, and visual traces can be presented through the existing image input \citep{zheng2025tracevla}. While these methods demonstrate the value of structured guidance, our interest is in reasoning supervision that can be shared across RFMs without additional architecture changes.

Natural language offers this opportunity because it can describe task conditions, spatial relationships, and intended outcomes without committing to a particular set of robot coordinates. It is already supported by the pretrained tokenizers and language components of many robot foundation models \citep{rt2,openvla,pi05}, and its annotations can be generated from existing demonstrations \citep{ecot}. 
Prior work uses natural language to describe fine-grained motions \citep{belkhale2024rt}, convey the current semantic subtask \citep{pi05}, and reason about plans, objects, and robot state before predicting actions \citep{mu2023embodiedgpt}. These methods show that language can provide useful intermediate supervision.

\section{Preliminaries and Problem Statement}
\label{sec:problem}

\tbm{\textbf{Robot task setting.}}
We consider a robot operating in a partially observed environment. A task is specified by an instruction $g \in \mathcal{G}$. At time $t$, the robot receives an observation $o_t \in \mathcal{O}$ and executes an action $a_t$ $\in \mathcal{A}$. We denote the interaction history by $h_t = (o_0, a_0, \ldots, a_{t-1}, o_t).$
To cover models that predict several actions at once, we write $\mathbf{a}_t = (a_t,\ldots,a_{t+H-1})$ for an action chunk of horizon $H$. A robot demonstration is a trajectory $\tau = (g,o_0,a_0,\ldots,o_T),$
and the training dataset is $\mathcal{D}=\{\tau_i\}_{i=1}^{N}$.
We treat a pretrained robot foundation model independently of its internal architecture. At each time step, it defines a conditional action distribution $\pi_{\theta_0}(\mathbf{a}_t \mid h_t,g),$
where $\theta_0$ denotes the pretrained parameters. Let
$S(\tau,g)\in[0,1]$ measure whether the resulting trajectory completes the task. The objective of the policy is to maximize
$J(\pi)=\mathbb{E}_{g,\tau\sim\pi}\left[S(\tau,g)\right].$

\tbm{\textbf{Reasoning as intermediate computation.}}
A standard policy predicts actions directly from the task and interaction history. We define \emph{reasoning} more generally as forming an intermediate representation $r_t \in \mathcal{R}$ that is formed before action prediction and carries information relevant to the current decision: $r_t \sim q_{\phi}(r_t \mid h_t,g), \text{ } \mathbf{a}_t \sim \pi_{\theta}(\mathbf{a}_t \mid h_t,g,r_t).$
The resulting action distribution is
$\pi_{\theta,\phi}(\mathbf{a}_t \mid h_t,g)=\int_{\mathcal{R}}\pi_{\theta}(\mathbf{a}_t \mid h_t,g,r_t)q_{\phi}(r_t \mid h_t,g)\,dr_t.$

\tbm{\textbf{Problem statement.}}
Given a pretrained robot foundation model \(\pi_{\theta_0}\) and a demonstration dataset \(\mathcal{D}\) containing observations and actions but no reasoning supervision, our goal is to learn a reasoned policy \(\pi_{\theta,\phi}\) that improves task success while retaining the model's existing control abilities. Crucially, this improvement should not require a new model architecture, a new foundation-scale pretraining stage, or the collection of another large robot dataset.

\section{The \GlauxLogo \name Framework}
\label{sec:glaux}

We now describe the components of \name: the reasoning representation (Sec.~\ref{sec:reasoning_representation}), the \tracename relabeling pipeline (Sec.~\ref{sec:glaux_trace}), and the fine-tuning recipe that connects reasoning to control (Sec.~\ref{sec:finetuning}).

\subsection{Action Grounded Causal Reasoning}
\label{sec:reasoning_representation}

We represent $r_t$ in natural language, reusing the model's pretrained language conditioning. We examine ECoT-style reasoning and semantic subtask supervision, adapting both formats to the same $\pi_{0.5}$ backbone and fine-tuning it with each \citep{ecot,pi05}.
\begin{wrapfigure}{r}{0.55\textwidth}
  \par\vspace{-8pt}
    \centering
    \includegraphics[
        width=\linewidth,
        trim=0 0 0 0,
        clip
    ]{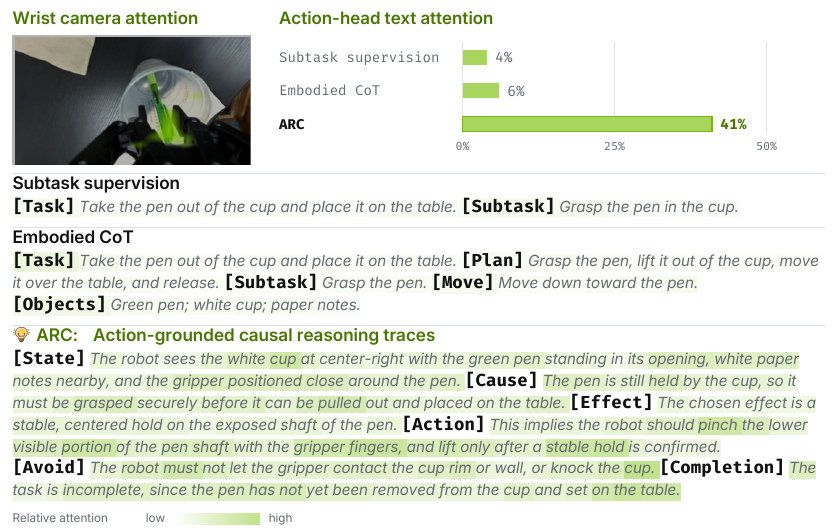}
        \par\vspace{-6pt}
    \caption{Reasoning traces and action-expert attention.}
    \label{fig:reasoning_representations}
        \par\vspace{-8pt}
\end{wrapfigure}
To determine whether these representations actually affect action prediction, we inspect how the action expert distributes attention over image and natural-language tokens in Fig.~\ref{fig:reasoning_representations}. Under both ECoT and subtask supervision, the action expert remains strongly focused on the visual observation and assigns little attention to the intermediate language. We treat attention as a diagnostic rather than proof of causal dependence, but the pattern exposes a plausible shortcut. Plans, object lists, motion directions, and subtask labels are often predictable from the same image and instruction already available to the model. The observed pattern is therefore consistent with shortcut learning, where the model can satisfy the language objective while continuing to predict actions from vision. 
Similar grounding failures have been reported in other robot foundation models, which can continue pursuing the original target even when the instruction specifies a different one \citep{liberoplus}. This suggests that the missing information is neither a longer scene description nor a finer decomposition of the task. An action is selected because of the change it is expected to produce in the world. Reasoning should make this relation explicit \citep{alpamayor1}. Following this principle, we develop an action-grounded causal representation for robot control.

For an action chunk $\mathbf{a}_t$, \tracename represents reasoning as
$r_t = [\mathrm{State},\allowbreak \mathrm{Cause},\allowbreak
\mathrm{Consequence},\allowbreak \mathrm{Effect},\allowbreak
\mathrm{Action},\allowbreak \mathrm{Avoid},\allowbreak
\mathrm{Completion}]_t$.
State and Cause identify the task-relevant facts and why they matter at the current step. Consequence reasons forward from those facts, while Effect selects the state change that should advance the task. Action connects that effect to the upcoming action chunk. Avoid records unwanted changes, and Completion preserves task progress across successive decisions. We call these action-grounded causal reasoning traces because they explain an action through the effect it is intended to cause, rather than describing the action in isolation.

When the same backbone is trained with \name traces, the action expert distributes substantially more attention to the natural-language reasoning, particularly to the spans that connect the current condition, intended effect, and resulting action. This does not by itself establish behavioral dependence. It does show that the \textit{trace now carries information used by the action expert} rather than remaining a redundant auxiliary target; Sec.~\ref{sec:experiments} examines how this change affects task performance.

\subsection{\name-Trace: Reasoning Supervision at Scale}
\label{sec:glaux_trace}

The representation above creates a supervision problem: existing robot datasets record \emph{what} the robot did, but not the reasoning that made each action appropriate. Action-grounded reasoning is therefore unavailable at the scale required to train robot foundation models. Collecting it alongside new demonstrations would also be costly. 
\begin{wrapfigure}{r}{0.5\textwidth}
    \centering
            \par\vspace{-6pt}
    \includegraphics[
        width=\linewidth,
        trim=0 20 0 0,
        clip
    ]{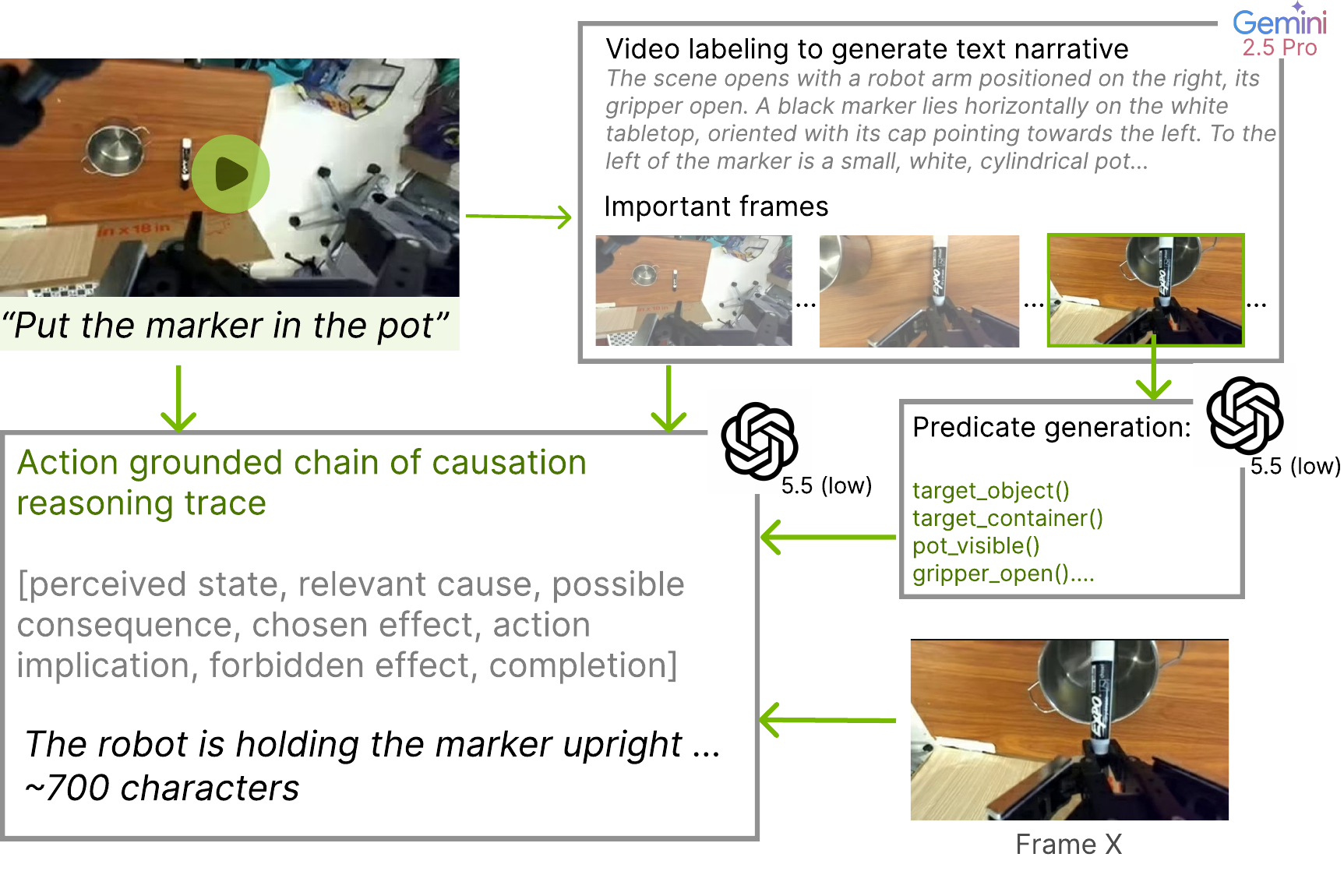}
  \par\vspace{-6pt}
\caption{Overview of the \GlauxLogo \name-Trace relabeling pipeline, illustrated with an example task.}    \label{arch_labeling}
  \par\vspace{-6pt}
\end{wrapfigure}
Our key observation is that the information needed to recover this supervision is already present in a successful demonstration. The task instruction specifies the goal, the video records how the scene changes, and the action sequence reveals how those changes were produced. Modern vision-language models can extract this structure and express it in natural language. We can therefore create reasoning supervision \emph{after} data collection, without gathering new robot trajectories or manually annotating individual actions.

We use this observation to construct \datasetname by automatically relabeling the DROID dataset with action-grounded causal reasoning. Figure~\ref{arch_labeling} summarizes the labeling pipeline. For each episode, we first generate a temporally grounded description of the task execution from the instruction and video. We then derive episode-specific state checks that capture the objects, relations, and task conditions needed to track progress through the demonstration. These checks provide structured evidence about what is true at each point in the episode and reduce the need for the final labeler to infer the entire task state from a single frame. Finally, at selected keyframes, a vision-language model combines the task instruction, episode description, current observations, and state checks to produce a reasoning trace aligned with the corresponding action chunk.
This process yields approximately $75\text{K}$ relabeled episodes and $1.2\text{M}$ action-aligned frames.
The full labeling prompts, sampling strategy, and filtering steps are provided in Appendix~\ref{app:labeling}.

\subsection{Fine-Tuning for Reasoning-Guided Control}
\label{sec:finetuning}

We fine-tune pretrained models to predict demonstrations conditioned on their aligned causal traces, observations, and task instructions. Traces reuse the base tokenizer and embeddings, and are processed by the language transformer with the task and visual context. The action-generation layers attend to these representations throughout denoising; token placement, projections, and masks depend on the architecture.

We retain each model's flow-matching objective, predicting actions for the VLA and actions plus future observations for the WAM. Let $\mathcal{K}$ index these modalities, with target $\mathbf{y}_t^k$ in normalized action or encoded-observation space. Its noisy counterpart is $\mathbf{x}_t^k=(1-s_k)\mathbf{y}_t^k+s_k\boldsymbol{\epsilon}^k$, with $\boldsymbol{\epsilon}^k\sim\mathcal{N}(0,I)$ and $s_k\in[0,1]$ sampled from the base noise schedule. The current observation $o_t$, instruction $g$, and trace $r_t$ remain uncorrupted. Collecting the noisy targets and noise levels as $\mathbf{x}_t$ and $\mathbf{s}$, we minimize
\begin{equation}
\mathcal{L}_{\mathrm{pred}}
=
\mathbb{E}\left[
\sum_{k\in\mathcal{K}}\alpha_k\,
\operatorname{MSE}_k\!\left(
v_{\theta}^{k}(\mathbf{x}_t,\mathbf{s}\mid o_t,g,r_t),
\boldsymbol{\epsilon}^{k}-\mathbf{y}_t^k
\right)\right].
\label{eq:glaux_prediction}
\end{equation}
Here, $v_{\theta}^{k}$ is the velocity predicted by model parameters $\theta$, and the expectation covers demonstrations, noise, and noise levels. Modality weights $\alpha_k$ and the masking, time weighting, and normalization in $\operatorname{MSE}_k$ follow the base recipe.

\tbm{\textbf{VLA Implementation.}}
\label{sec:vla_finetuning}
We use $\pi_{0.5}$-DROID \citep{pi05}, with a PaliGemma backbone comprising a SigLIP vision encoder and Gemma language model, and a separate 300M-parameter action expert. A two-layer encoder maps PaliGemma trace embeddings to tokens appended to the image-and-instruction context. The trace attends to this context, and the action expert attends to both; all components are jointly fine-tuned. Here, $\mathbf{y}_t^{\mathrm{act}}=\mathbf{a}_t$ is the demonstrated action chunk.

With factual supervision alone, predictions change little under conflicting traces for the same observation. To discourage this shortcut, we replace approximately $5\%$ of traces with counterfactuals $r_t^{-}$ contradicting the demonstrated action and exclude these examples from $\mathcal{L}_{\mathrm{pred}}$. Setting $s_{\mathrm{act}}=1$ and writing $\boldsymbol{\epsilon}=\boldsymbol{\epsilon}^{\mathrm{act}}$, the one-step estimate is $\widehat{\mathbf{a}}_t^{-}=\boldsymbol{\epsilon}-v_{\theta}^{\mathrm{act}}(\boldsymbol{\epsilon},1\mid o_t,g,r_t^{-})$. We penalize its proximity to the demonstration:
\begin{equation}
\begin{aligned}
\mathcal{L}_{\mathrm{cf}}
&=\mathbb{E}\left[
\max\left(0,m-\operatorname{MAE}_{\mathrm{real}}
(\widehat{\mathbf{a}}_t^{-},\mathbf{a}_t)\right)^2
\right], \quad
\mathcal{L}_{\mathrm{VLA}}
&=\mathcal{L}_{\mathrm{pred}}+
\lambda_{\mathrm{cf}}\mathcal{L}_{\mathrm{cf}}.
\end{aligned}
\label{eq:glaux_vla}
\end{equation}
The expectation in $\mathcal{L}_{\mathrm{cf}}$ covers counterfactual samples and action noise; $\mathcal{L}_{\mathrm{pred}}$ uses only factual samples. The margin $m>0$ is in normalized action space, $\lambda_{\mathrm{cf}}$ weights the penalty, and $\operatorname{MAE}_{\mathrm{real}}$ averages over the horizon and eight non-padding action coordinates. This supplies negative supervision without alternative-action demonstrations. Implementation details appear in Appendix~\ref{app:finetuning}.

\tbm{\textbf{WAM Implementation.}}
\label{sec:wam_finetuning}
Cosmos3-Nano-Policy \citep{cosmos3} combines a vision-language reasoner and diffusion generator with separate parameters connected through attention at every layer. The generator attends to both token sets; the reasoner attends only to its own, allowing trace conditioning without architectural changes. Each window contains one observed frame and 32 future frames with actions, paired with the first frame's trace. We jointly fine-tune both components with
\begin{equation}
\begin{aligned}
\mathcal{L}_{\mathrm{WAM}}
&=\mathcal{L}_{\mathrm{pred}}+
\lambda_{\mathrm{trace}}\mathcal{L}_{\mathrm{trace}},\quad
\mathcal{L}_{\mathrm{trace}}
&=-\mathbb{E}\left[
\sum_{j=1}^{|r_t|}
\log q_{\phi}(r_{t,j}\mid o_t,g,r_{t,<j})
\right].
\end{aligned}
\label{eq:glaux_wam}
\end{equation}
Here, $q_{\phi}$ is the reasoner's next-token distribution with parameters $\phi$, $|r_t|$ is the trace length, and $r_{t,j}$ and $r_{t,<j}$ denote token $j$ and its prefix. The coefficient $\lambda_{\mathrm{trace}}$ weights trace supervision, whose expectation is over labeled windows. The generator conditions on ground-truth traces, and prediction-loss gradients also update the reasoner. Cosmos3's larger reasoner (8B versus $\pi_{0.5}$'s 2B) already produces distinct action distributions under conflicting traces. Adding $\mathcal{L}_{\mathrm{cf}}$ yields no measurable gain, so we omit it. Supporting ablations and training details appear in Appendix~\ref{app:cosmos}.

\tbm{\textbf{Inference.}}
At inference, we freeze the \name-fine-tuned models and use an external VLM to generate reasoning traces from the task instruction and current camera observations, refreshing them at a selected frequency during execution. The VLA's vision-language backbone and the WAM's reasoner encode these traces into representations that condition the action expert and generator, respectively. We use external generation because the built-in language components are less reliable at generating these traces than at using them for control (Section~\ref{sec:key_findings}). Separately, \name{} maintains comparable task performance across the external VLMs evaluated. Since the traces are natural language, their generation can benefit directly from standard LLM inference optimizations. Appendix~\ref{app:inference} details the refresh schedules and serving optimizations, including a near-real-time VLA deployment with $\sim 10\,\mathrm{ms}$ of added control-loop overhead.

\section{Experiments}
\label{sec:experiments}
We evaluate $\pi_{0.5}$+\name (VLA) and Cosmos3-Nano+\name (WAM) (both finetuned on \datasetname) zero-shot on RoboLab-120 \citep{robolab}, MolmoSpaces \citep{kim2026molmospaces}, and RoboLab-Reasoning-50. We  additionally evaluate \name-$\pi_{0.5}$ on paired real-world tasks without further fine-tuning. All simulation and hardware experiments use the DROID embodiment. We use Qwen3.6-35B-A3B as the default external VLM (ablated in section~\ref{sec:key_findings}).

\subsection{Quantitative Results: Simulation and Hardware Benchmarks}
\label{sec:quantitative_sim}

\textbf{General task performance.}
Tables~\ref{tab:arc_robolab120} and~\ref{tab:arc_molmospaces} highlight task success for \name-variants, its base models, on RoboLab and MolmoSpaces. Across both benchmarks, \name substantially improves both base models, enabling the comparatively weaker $\pi_{0.5}$ to outperform stronger pretrained policies while further improving Cosmos3-Nano-Policy. Importantly, these gains hold without benchmark-specific training, demonstrating that the same reasoning recipe improves general manipulation across different architectures and starting levels of performance.
Table breakdowns, multi-seed runs, evaluation settings and baseline details are provided in Appendix~\ref{detailed-app}.
\begingroup
\definecolor{arcbmDeep}{HTML}{2F4D00}
\definecolor{arcbmMuted}{HTML}{92989F}
\definecolor{arcbmTint}{HTML}{F1F8E6}
\definecolor{arcbmAccent}{HTML}{A8D55E}
\definecolor{arcbmRule}{HTML}{DCE2E7}
\definecolor{arcbmGain}{HTML}{4A7500}
\definecolor{arcbmNvidia}{HTML}{76B900}
\newcommand{\arcbmarclogo}{%
  \raisebox{-0.1em}{\includegraphics[height=1em]{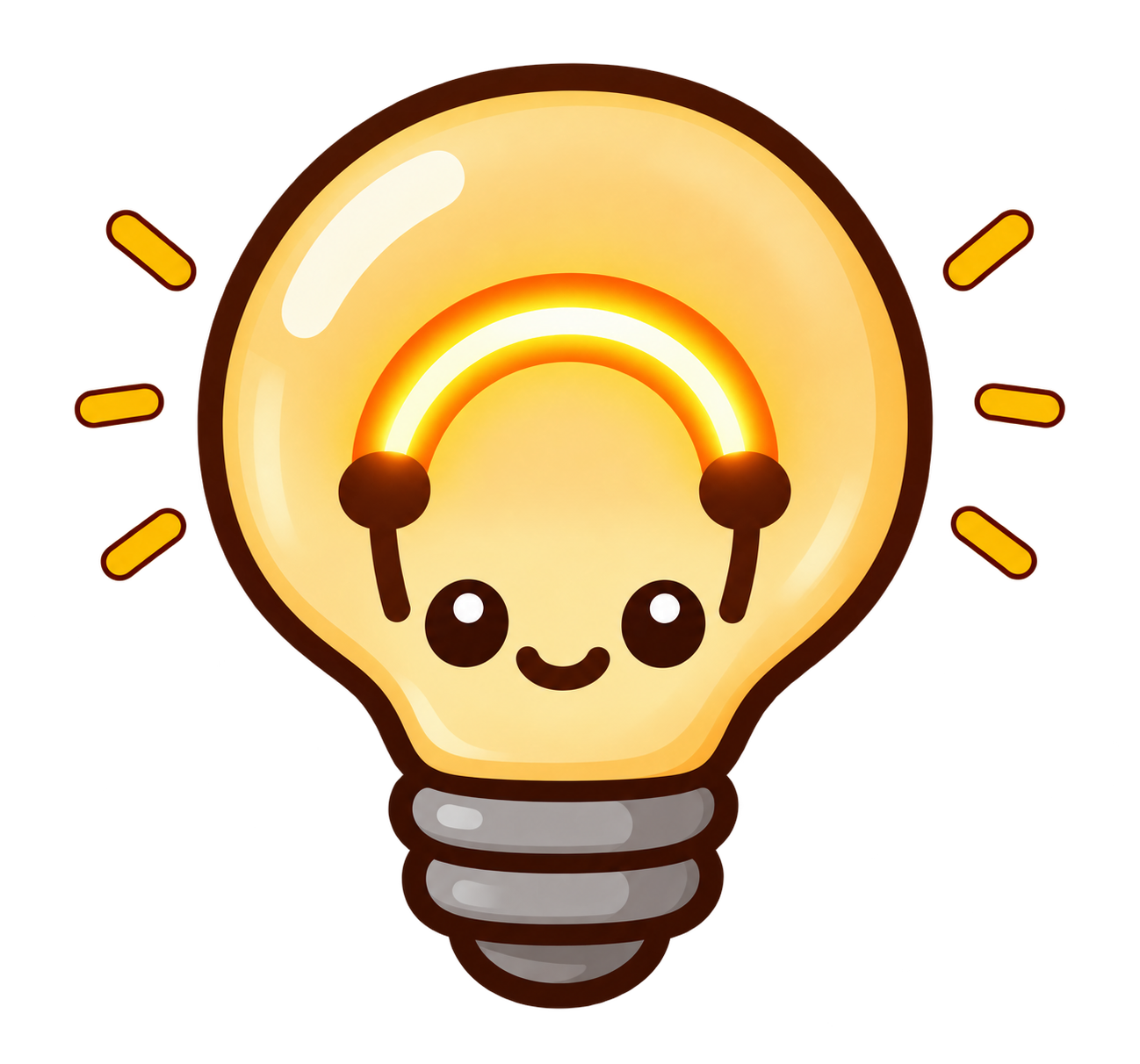}}}
\newcommand{\arcbmarc}{\arcbmarclogo\,ARC}
\newcommand{\arcbmpi}{\ensuremath{\pi_{0.5}}}
\newcommand{\arcbmpifast}{\ensuremath{\pi_{0}}-FAST}
\newcommand{\arcbmbold}[1]{\textcolor{arcbmDeep}{\bfseries\boldmath #1}}
\newcommand{\arcbmline}{\specialrule{0.25pt}{0pt}{0pt}}
\newcommand{\arcbmaccent}{%
  \arrayrulecolor{arcbmAccent}\specialrule{0.65pt}{0pt}{0pt}%
  \arrayrulecolor{arcbmRule}}
\ifdefined\arcbmgap\else\newlength{\arcbmgap}\fi
\settowidth{\arcbmgap}{\hspace{3pt}{\fontencoding{T1}\fontfamily{phv}\fontsize{6.2}{7}\selectfont\bfseries +00.0}}
\ifdefined\arcbmmolmogap\else\newlength{\arcbmmolmogap}\fi
\setlength{\arcbmmolmogap}{\arcbmgap}
\newcommand{\arcbmrate}[1]{#1\%\makebox[\arcbmgap]{}}
\newcommand{\arcbmgain}[2]{%
  \arcbmbold{#1\%}\makebox[\arcbmgap][l]{\hspace{3pt}%
  {\color{arcbmGain}\fontsize{6.2}{7}\selectfont\bfseries +#2}}}
\newcommand{\arcbmpending}{%
  {\fontsize{6.2}{7}\selectfont\slshape running}\makebox[\arcbmgap]{}}
\newcommand{\arcbmmrate}[1]{#1\%\makebox[\arcbmmolmogap]{}}
\newcommand{\arcbmmgain}[2]{%
  \arcbmbold{#1\%}\makebox[\arcbmmolmogap][l]{\hspace{3pt}%
  {\color{arcbmGain}\fontsize{6.2}{7}\selectfont\bfseries +#2}}}
\newcommand{\arcbmmrunning}[1]{%
  \arcbmbold{#1\%}\makebox[\arcbmmolmogap][l]{\hspace{3pt}%
  {\color{arcbmMuted}\fontsize{6.2}{7}\selectfont\slshape (running)}}}
\newcommand{\arcbmrobot}{%
  \raisebox{-2.5pt}{\includegraphics[height=14pt,trim=20 8 21 7,clip]{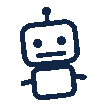}}}
\newcommand{\arcbmroboheading}[1]{%
  \multicolumn{3}{@{}l@{}}{\rule{0pt}{19pt}\hspace{2pt}%
  \arcbmrobot\hspace{3pt}%
  {\fontsize{10.5}{12}\selectfont\bfseries
   \textcolor{arcbmNvidia}{RoboLab-120}\textcolor{arcbmDeep}{: #1}}}\\[3pt]
  {\fontsize{7}{8}\selectfont\#} &%
  {\fontsize{7}{8}\selectfont Policy} &%
  {\fontsize{7}{8}\selectfont SR}\makebox[\arcbmgap]{}\\
  \arcbmaccent}
\newcommand{\arcbmmolmoheading}{%
  \multicolumn{4}{@{}l@{}}{\rule{0pt}{19pt}\hspace{2pt}%
  \raisebox{-1pt}{\includegraphics[height=12pt,trim=0 21 8 52,clip]{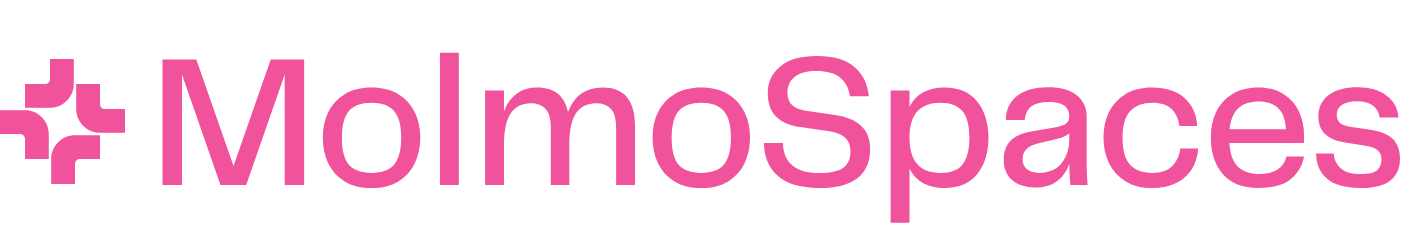}}}\\[3pt]
  {\fontsize{7}{8}\selectfont\#} &%
  {\fontsize{7}{8}\selectfont Policy} &%
  {\fontsize{7}{8}\selectfont SR}\makebox[\arcbmmolmogap]{} &%
  {\fontsize{7}{8}\selectfont\# tasks}\\
  \arcbmaccent}
\newcommand{\arcbmbase}[3]{%
  \textcolor{black}{#1} & \textcolor{black}{#2} &%
  \textcolor{black}{\arcbmrate{#3}}\\}
\newcommand{\arcbmmbase}[4]{%
  \textcolor{black}{#1} & \textcolor{black}{#2} &%
  \textcolor{black}{\arcbmmrate{#3}} & \textcolor{black}{#4}\\}
\newcommand{\arcbmsetup}{%
\fontencoding{T1}\fontfamily{phv}\fontsize{9}{10.8}\selectfont
\color{arcbmMuted}
\setlength{\tabcolsep}{2.5pt}
\renewcommand{\arraystretch}{1.55}
\setlength{\aboverulesep}{0pt}
\setlength{\belowrulesep}{0pt}
\arrayrulecolor{arcbmRule}
}
\ifdefined\arcbmRoboBox\else\newsavebox{\arcbmRoboBox}\fi
\ifdefined\arcbmMolmoBox\else\newsavebox{\arcbmMolmoBox}\fi

\begin{table}[H]
\centering
\begin{minipage}[t]{0.7133\linewidth}
\vspace{0pt}
\centering
\begingroup
\arcbmsetup
\sbox{\arcbmRoboBox}{%
\begin{tabular}[t]{r l r}
\arcbmroboheading{Vague}
\rowcolor{arcbmTint}
\arcbmbold{1} & \arcbmbold{\arcbmpi{} + \arcbmarc} & \arcbmgain{45.1}{29.8}\\
\rowcolor{arcbmTint}
\arcbmbold{2} & \arcbmbold{Cosmos3-Nano + \arcbmarc} & \arcbmgain{44.9}{24.3}\\
\arcbmline
3 & Atomic-WAM & \arcbmrate{31.6}\\
4 & BiMind v0.1 & \arcbmrate{30.2}\\
5 & VoLo & \arcbmrate{29.5}\\
\arcbmbase{6}{Cosmos3-Nano-Policy}{20.6}
7 & Cosmos3-Edge-Policy & \arcbmrate{15.4}\\
\arcbmbase{8}{\arcbmpi}{15.2}
9 & DreamZero & \arcbmrate{14.9}\\
10 & \arcbmpifast & \arcbmrate{9.2}\\
11 & GR00T N1.6 & \arcbmrate{5.4}\\
\arcbmline
\end{tabular}%
\hspace{5pt}%
\begin{tabular}[t]{r l r}
\arcbmroboheading{Default}
\rowcolor{arcbmTint}
\arcbmbold{1} & \arcbmbold{Cosmos3-Nano + \arcbmarc} & \arcbmgain{48.8}{12.0}\\
\rowcolor{arcbmTint}
\arcbmbold{2} & \arcbmbold{\arcbmpi{} + \arcbmarc} & \arcbmgain{45.3}{17.3}\\
\arcbmline
3 & FLUX 3 Action & \arcbmrate{42.9}\\
4 & HiDream-O1-Embodied & \arcbmrate{39.9}\\
5 & Atomic-WAM & \arcbmrate{39.6}\\
6 & OASIS WAM & \arcbmrate{39.0}\\
\arcbmbase{7}{Cosmos3-Nano-Policy}{36.8}
8 & Phoenix & \arcbmrate{34.4}\\
9 & BiMind v0.1 & \arcbmrate{33.3}\\
10 & VoLo & \arcbmrate{28.2}\\
\arcbmbase{11}{\arcbmpi}{28.0}
\arcbmline
\end{tabular}%
\hspace{5pt}%
\begin{tabular}[t]{r l r}
\arcbmroboheading{Specific}
\rowcolor{arcbmTint}
\arcbmbold{1} & \arcbmbold{Cosmos3-Nano + \arcbmarc} & \arcbmgain{51.2}{11.5}\\
\rowcolor{arcbmTint}
\arcbmbold{2} & \arcbmbold{\arcbmpi{} + \arcbmarc} & \arcbmgain{45.0}{16.9}\\
\arcbmline
\arcbmbase{3}{Cosmos3-Nano-Policy}{39.7}
4 & VoLo & \arcbmrate{31.4}\\
5 & BiMind v0.1 & \arcbmrate{30.6}\\
6 & Cosmos3-Edge-Policy & \arcbmrate{28.8}\\
\arcbmbase{7}{\arcbmpi}{28.1}
8 & DreamZero & \arcbmrate{23.9}\\
9 & \arcbmpifast & \arcbmrate{14.9}\\
10 & paligemma-binning & \arcbmrate{5.5}\\
11 & GR00T N1.6 & \arcbmrate{5.3}\\
\arcbmline
\end{tabular}%
}
\typeout{ARC-ROBO-NATURAL=\the\wd\arcbmRoboBox}
\resizebox{\linewidth}{!}{\usebox{\arcbmRoboBox}}\par
\endgroup
\caption{RoboLab-120 benchmark results. SR is success rate; +values give gains over the corresponding base policy in p.p.}
\label{tab:arc_robolab120}
\end{minipage}\hfill%
\begin{minipage}[t]{0.2687\linewidth}
\vspace{0pt}
\centering
\begingroup
\arcbmsetup
\sbox{\arcbmMolmoBox}{%
\begin{tabular}[t]{r l r r}
\arcbmmolmoheading
\rowcolor{arcbmTint}
\arcbmbold{1} & \arcbmbold{Cosmos3-Nano + \arcbmarc} & \arcbmmgain{57.6}{18.6} & \arcbmbold{9/9}\\
\arcbmline
2 & ZoeInside-VLA & \arcbmmrate{49.5} & 9/9\\
3 & Phoenix & \arcbmmrate{48.0} & 9/9\\
\rowcolor{arcbmTint}
\arcbmbold{4} & \arcbmbold{\arcbmpi{} + \arcbmarc} & \arcbmmgain{45.0}{27.2} & \arcbmbold{9/9}\\
\arcbmline
\arcbmmbase{5}{Cosmos3-Nano-Policy}{39.0}{9/9}
6 & WALL-OSS-0.5 & \arcbmmrate{38.5} & 9/9\\
7 & TiPToP & \arcbmmrate{34.7} & 7/9\\
8 & PRTS-Droid & \arcbmmrate{27.8} & 9/9\\
9 & MolmoAct2 DROID & \arcbmmrate{25.9} & 9/9\\
10 & Psi-R2 & \arcbmmrate{22.8} & 9/9\\
\arcbmmbase{11}{\arcbmpi{} DROID}{17.8}{9/9}
\arcbmline
\end{tabular}%
}
\typeout{ARC-MOLMO-NATURAL=\the\wd\arcbmMolmoBox}
\resizebox{\linewidth}{!}{\usebox{\arcbmMolmoBox}}\par
\endgroup
\caption{MolmoSpaces benchmark results.}
\label{tab:arc_molmospaces}
\end{minipage}
\end{table}
\endgroup

\edef\arcFullTableWidth{\the\linewidth}

\begin{wraptable}{r}{0.3\linewidth}
          \par\vspace{-6pt}
  \centering
  \begingroup
\fontencoding{T1}\fontfamily{phv}\fontsize{6}{7.5}\selectfont
\definecolor{roDeep}{HTML}{2F4D00}
\definecolor{roTint}{HTML}{F1F8E6}
\definecolor{roAccent}{HTML}{A8D55E}
\definecolor{roRule}{HTML}{DCE2E7}
\definecolor{roGreen}{HTML}{76B900}
\definecolor{roGainGreen}{HTML}{4A7500}
\color{black}
\setlength{\tabcolsep}{1.5pt}
\renewcommand{\arraystretch}{1.2}
\setlength{\aboverulesep}{0pt}
\setlength{\belowrulesep}{0pt}
\arrayrulecolor{roRule}
\edef\roPanelWidth{\the\linewidth}

\def\roPi{\ensuremath{\pi_{0.5}}}
\def\roARC{%
  \raisebox{-0.15em}{\includegraphics[height=1.05em]{figures/arc.png}}\,ARC}
\def\roBold#1{\textcolor{roDeep}{\bfseries\boldmath #1}}
\def\roRate#1{%
  \makebox[18pt][r]{#1\%}\makebox[17pt]{}}
\def\roGain#1#2{%
  \makebox[18pt][r]{\roBold{#1\%}}%
  \makebox[17pt][l]{\hspace{1.5pt}%
    \textcolor{roGainGreen}{\fontsize{5.2}{6.2}\selectfont\bfseries +#2}}}
\def\roSRHead{%
  \makebox[18pt][r]{\roBold{SR}}\makebox[17pt]{}}
\def\roLine{\specialrule{0.25pt}{0pt}{0pt}}
\def\roAccentLine{%
  \arrayrulecolor{roAccent}\specialrule{0.65pt}{0pt}{0pt}%
  \arrayrulecolor{roRule}}
\def\roTitle#1{%
  \noindent
  \raisebox{-1pt}{\includegraphics[height=8pt,trim=20 8 21 7,clip]{figures/robolab.png}}%
  \hspace{2pt}{\color{roGreen}\bfseries\fontsize{6.5}{7.8}\selectfont RoboLab-#1}%
  \par\nobreak\vspace{3pt}}

\noindent
\begin{minipage}[t]{\roPanelWidth}
\vspace{0pt}
\roTitle{Reasoning-50}
\begin{tabular}{%
  >{\raggedright\arraybackslash}m{\dimexpr\linewidth-4\tabcolsep-35pt\relax}
  >{\raggedleft\arraybackslash}m{35pt}}
\roBold{Policy} & \roSRHead\\
\roAccentLine
\rowcolor{roTint}
\roBold{Cosmos3-Nano + \roARC} & \roGain{61.0}{50.0}\\
\roLine
\rowcolor{roTint}
\roBold{\roPi{} + \roARC} & \roGain{57.0}{46.8}\\
\roLine
\roPi{} + explicit subtasks & \roRate{21.0}\\
\roLine
Cosmos3-Nano-Policy & \roRate{11.0}\\
\roLine
\roPi & \roRate{10.2}\\
\roLine
\end{tabular}
\end{minipage}
\par\vspace{6pt}
\noindent
\begin{minipage}[t]{\roPanelWidth}
\vspace{0pt}
\roTitle{Reasoning-Hardware}
\begin{tabular}{%
  >{\raggedright\arraybackslash}m{\dimexpr\linewidth-4\tabcolsep-35pt\relax}
  >{\raggedleft\arraybackslash}m{35pt}}
\roBold{Policy} & \roSRHead\\
\roAccentLine
\rowcolor{roTint}
\roBold{\roPi{} + \roARC} & \roGain{91.7}{82.2}\\
\roLine
Cosmos3-Nano-Policy & \roRate{11.9}\\
\roLine
\roPi & \roRate{9.5}\\
\roLine
\end{tabular}
\end{minipage}
\par
\endgroup

  \caption{Reasoning-50 and
    Reasoning-Hardware results.}
  \label{tab:arc_reasoning}
          \par\vspace{-8pt}
\end{wraptable}

\textbf{Reasoning-focused evaluation.}
RoboLab and MolmoSpaces target general task performance. To evaluate reasoning more directly, we introduce RoboLab-Reasoning-50, a reasoning-focused benchmark built on RoboLab. This complements the broader evaluations by testing \name on tasks specifically designed to require reasoning during execution. 
Specifically, this benchmark evaluates four major capabilities that require reasoning: 1) Contextual tasks, requiring visual and semantic understanding; 2) Long-Horizon, requiring memory; 3) Discovery, requiring exploration via scene interaction; 4) Negation, requiring understanding exclusions. 
Table~\ref{tab:arc_reasoning} compares \name and the baselines on this suite. The benchmark construction, individual tasks, and evaluation protocol are described in Appendix~\ref{app:robolab_reasoning50}. Beyond general manipulation, \name substantially improves both models on tasks requiring them to reason,  infer unstated steps and complete long-horizon plans.

\textbf{Hardware Evaluation.}
We deploy $\pi_{0.5}$+\name{} alongside $\pi_{0.5}$ and Cosmos3-Nano-Policy on the DROID hardware setup using the same checkpoints evaluated in simulation. We recreate 28 tasks from RoboLab-Reasoning-50 on hardware. Details of the comparison policies are provided in Appendix~\ref{app:hardware}.
Table~\ref{tab:arc_reasoning} reports task success across categories and the corresponding simulation results. These paired evaluations test whether the improvements from reasoning persist during physical execution. The gains observed in simulation carry over to physical execution, with $\pi_{0.5}$+\name substantially improving $\pi_{0.5}$'s task success on paired hardware tasks without hardware-specific fine-tuning.
Category definitions, hardware specifications and evaluation procedures appear in Appendix~\ref{app:hardware}.

\begin{figure}[H]
    \centering
    \includegraphics[width=1\linewidth,  trim={0pt 0pt 0pt 0pt},
    clip]{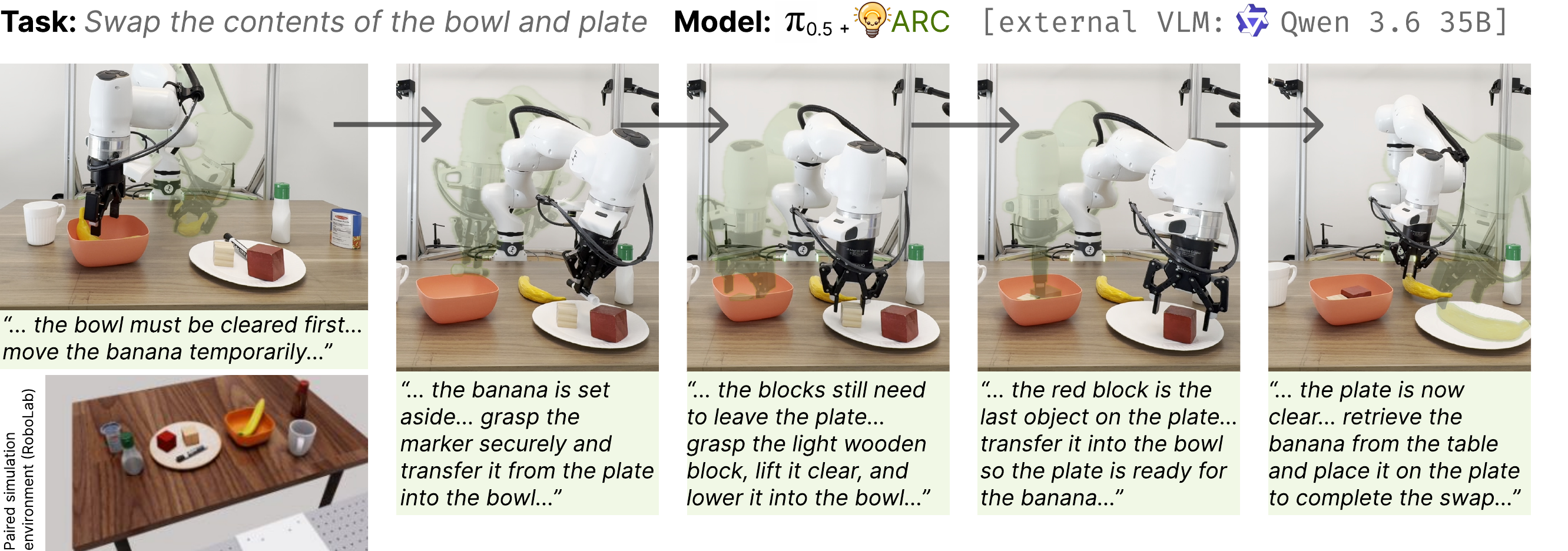}
\caption{Hardware rollout of $\pi_{0.5}$+\name{} on RoboLab-Reasoning-Hardware. The VLM dynamically refreshes the reasoning traces ($\sim 1\,\mathrm{Hz}$ in this episode), while robot control runs at $15\,\mathrm{Hz}$.}
    \label{fig:placeholder}
\end{figure}

\subsection{Key Findings and Ablations}
\label{sec:key_findings}

Tables~\ref{tab:arc_robolab120}, \ref{tab:arc_molmospaces} and~\ref{tab:arc_reasoning} support the central finding of this work that stronger grounding of causal language leads to stronger overall performance.
The improvement extends across both model families. \tbm{\nametwo{} lifts comparatively weaker models such as $\pi_{0.5}$ above unmodified WAM baselines, while substantially improving already strong models such as Cosmos3-Nano-Policy.} These gains extend beyond task success to robustness and efficiency. We identify four key findings (KF) that indicate \name as a useful recipe for uplifting existing RFM performance.

\begin{wrapfigure}{r}{0.48\textwidth}
  \centering
  \includegraphics[
    width=\linewidth,
  ]{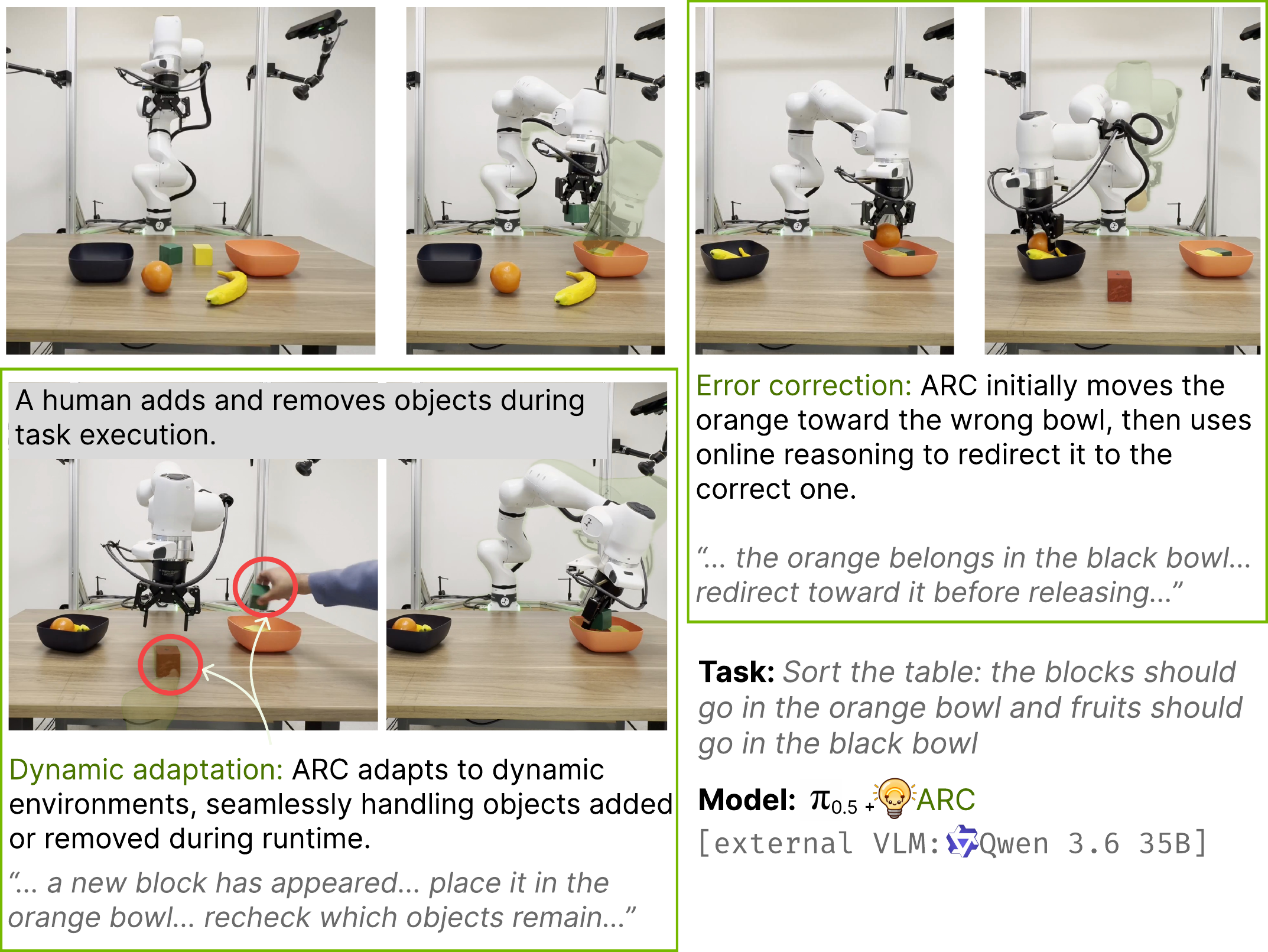}
  \par\vspace{-6pt}
\caption{Illustrative hardware rollout on a long-horizon, reasoning-intensive task in a dynamic environment.}  \label{hwreal}
  \par\vspace{-6pt}
\end{wrapfigure}
\tbm{\textbf{(KF\#1)} Reasoning improves execution across long-horizon, dynamic, and reasoning-intensive tasks.}
The reasoning-focused evaluations and hardware rollouts reveal additional benefits (Tables~\ref{tab:arc_reasoning}; Fig.~\ref{hwreal}). In long-horizon tasks, \name tracks progress and maintains the ordering of the remaining steps. When the scene changes during execution, updated reasoning redirects the robot toward what is currently needed. Reasoning-intensive tasks expose another benefit: the robot identifies prerequisites absent from the instruction and takes the intermediate steps needed to complete the task. These behaviors are accompanied by stronger error recovery, with failed actions followed by corrected attempts. Our interpretation is that explicitly connecting the current state to the intended effect supports several forms of decision-making through the same representation: sequencing actions, discovering necessary steps, responding to changes, and correcting mistakes.

\begin{wrapfigure}{r}{0.48\textwidth}
    \par\vspace{-10pt}
  \centering
  \includegraphics[
    width=\linewidth,
  ]{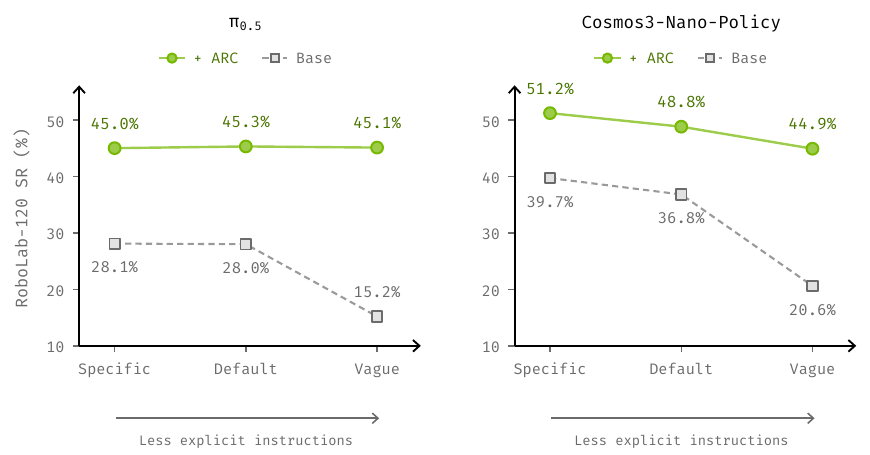}
    \par\vspace{-6pt}
  \caption{Instruction robustness on RoboLab-120.}
  \label{insp}
    \par\vspace{-6pt}
\end{wrapfigure}
\tbm{\textbf{(KF\#2)} Performance becomes nearly invariant to instruction specificity.}
On RoboLab-120, policies finetuned using \name maintains comparable success across vague, default, and specific instructions, while the base policies deteriorate as instructions become less explicit (Fig.~\ref{insp}). This naturally suggests that \textit{reasoning reduces the amount of detail the user must supply}. The trace interprets the instruction in the context of the scene and identifies the intended effect of the next action, giving the controller a more specific target even when the original instruction is underspecified.

\begin{wrapfigure}{r}{0.48\textwidth}
  \centering
  \includegraphics[
    width=\linewidth,
    trim={3pt 5pt 3pt 3pt},
    clip
  ]{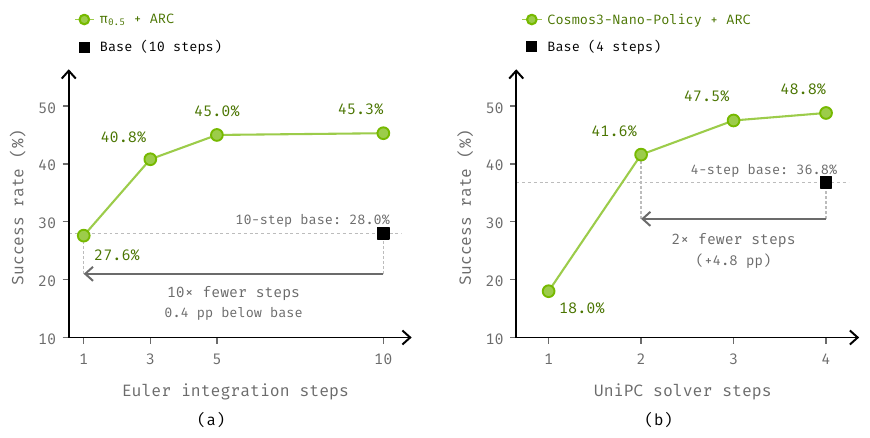}
    \par\vspace{-6pt}
\caption{Inference-time efficiency on RoboLab-120 (default).}
  \par\vspace{-6pt}
  \label{kf3}
\end{wrapfigure}
\tbm{\textbf{(KF\#3)} Reasoning buys inference-time speed-up.}
On RoboLab-120, $\pi_{0.5}$+\name matches its ten-step base policy with a single Euler step ($10\times$ improvement), while Cosmos3-Nano+\name matches its four-step base policy with two UniPC steps ($2\times$ improvement)(Fig.~\ref{kf3}). We hypothesize that \textit{specifying the intended effect narrows the set of plausible actions, reducing multimodality and simplifying the flow each model must integrate.} Coarse integration can then suffice because more of the decision has already been resolved through reasoning. Cosmos3-Nano still loses performance at one step, consistent with UniPC losing the multistep history needed for higher-order correction.

\begin{wrapfigure}{r}{0.48\textwidth}
    \par\vspace{-6pt}
  \centering
  \includegraphics[
    width=\linewidth,
    trim={6pt 6pt 0pt 0pt},
    clip
  ]{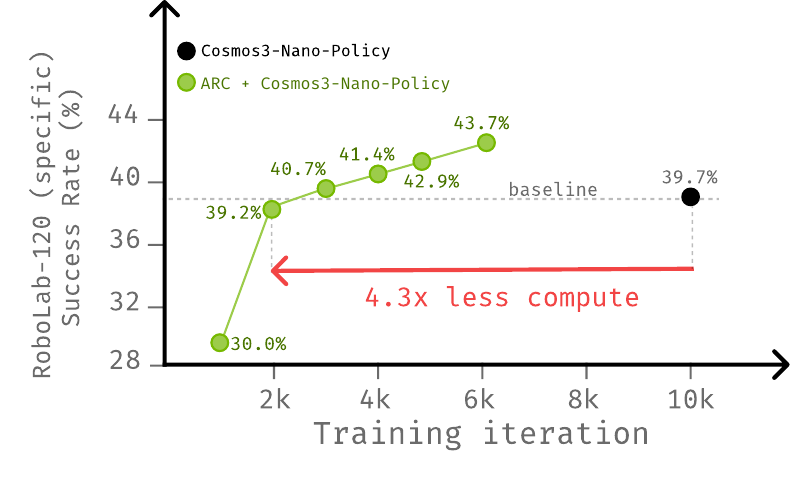}
    \par\vspace{-6pt}
  \caption{Training efficiency on RoboLab-120 (specific) for Cosmos3-Nano-Policy with and without \name.}
  \label{fig:training_scaling}
    \par\vspace{-6pt}
\end{wrapfigure}

\tbm{\textbf{(KF\#4)} Reasoning reaches target performance with fewer training updates.}
Figure~\ref{fig:training_scaling} shows that \name reaches the instruction-only baseline's 10K-iteration performance with approximately $4.3\times$ fewer updates, and continues to improve with further training. We conduct this study on Cosmos3-Nano-Policy using its open training implementation. Our interpretation is that \textit{causal supervision makes each demonstration more informative.} The model receives the reason for the demonstrated action alongside the action itself, reducing what it must infer from repeated observation--action examples alone. Reasoning thus improves how efficiently training produces useful behavior, in addition to improving the final policy.

\begin{figure}[H]
      \par\vspace{-6pt}
    \centering
    \includegraphics[width=0.9\linewidth]{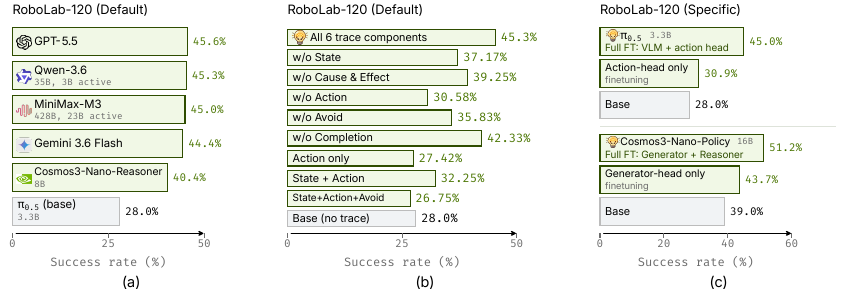}
      \par\vspace{-6pt}
    \caption{Key ablations of ARC on RoboLab-120.}
    \label{fig:abl}
\end{figure}

\textbf{Key ablations.}
We conduct three key design ablations on RoboLab-120. \textbf{(AB\#1}) First, we vary the external VLM while keeping the \name-fine-tuned $\pi_{0.5}$ checkpoint frozen (Fig.~\ref{fig:abl} (a)). Performance remains comparable across the VLMs evaluated, showing that \name is not tied to a particular trace generator. 
\textbf{(AB\#2)} Second, we separately fine-tune $\pi_{0.5}$ for each trace variant (Fig.~\ref{fig:abl} (b)). The results points to two complementary roles within the trace: The causal account identifies which state change matters and why, while the action implication connects that change to a physical step. Without the action implication, the controller must infer how to realize the intended effect; with the action implication alone, it receives a command without the causal context that makes it appropriate. Forbidden effects likewise constrain an intended behavior but do not specify one, explaining why they help within a complete causal account but not necessarily in isolation.
\textbf{(AB\#3)} Finally, we compare full fine-tuning with action-expert-only training in the VLA and generator-only training in the WAM (Fig.~\ref{fig:abl}(c)). Our interpretation is that pretrained language features do not expose causal distinctions in a form the controller can readily use. Action-head-only training leaves these features fixed and expects the controller to recover the relevant distinctions from them. Full fine-tuning allows prediction errors to reshape the features themselves, teaching the backbone which conditions and intended effects must be distinguished because they require different actions.

\section{Conclusion}

We presented \GlauxLogo \name, a reasoning recipe that substantially improves zero-shot task execution in pretrained RFMs without collecting new robot demonstrations or repeating foundation-scale training. Applied to $\pi_{0.5}$ and Cosmos3-Nano-Policy, \name{} establishes a new state of the art on RoboLab-120 and MolmoSpaces, improves robustness to ambiguous and underspecified instructions, and increases training and inference efficiency. Promising future directions include policies that reliably generate their own causal traces and extending the recipe across embodiments. \tbm{In simpler terms, we discover that RFMs do not need to be rebuilt to reason. Grounded causal language is a cheap, high-leverage axis for scaling them.}

\newpage
\section*{AI Use Statement}
We used vision-language models to automatically generate action-grounded causal annotations for ARC-Trace and, during evaluation, to generate reasoning traces from task instructions and camera observations. The models and prompting procedures are described in the methodology and appendix. Large language models also assisted with manuscript drafting and editing. The authors take responsibility for the final manuscript, reported results, and all AI-assisted content.

\section*{Reproducibility statement}
We describe ARC's reasoning representation and training objectives in Sections~\ref{sec:reasoning_representation} and~\ref{sec:finetuning}. Appendix~\ref{app:labeling} details the ARC-Trace construction pipeline and labeling prompts. Model-specific training configurations and hyperparameters are provided in Appendices~\ref{app:finetuning} and~\ref{app:cosmos}, while inference prompts and deployment procedures are documented in Appendices~\ref{app:external_vlm} and~\ref{app:inference}. The evaluation protocols, benchmark construction, baseline implementations, and hardware setup are described in Appendices \ref{app:robolab_reasoning50}, \ref{app:policies}, and~\ref{app:hardware}. Rollout videos are available on our fully anonymized project website at \url{https://arc-robot-reasoning.github.io}. The models, code, and ARC-Trace dataset will be released after the review period.

\bibliography{iclr2027_conference}
\bibliographystyle{bibstyle}
\newpage

\clearpage

\appendix

\section*{Appendix}
This appendix provides detailed recipes for data labeling, training, and inference, together with hyperparameters, prompts, additional qualitative results, and a discussion of limitations. All experiment videos are available on our fully anonymized project website at \url{https://arc-robot-reasoning.github.io}. The models, code, and ARC-Trace dataset will be released after the review period.

\startcontents[appendix]
\begingroup
\color{arcLink}
\ifdefined\hypersetup
  \hypersetup{linkcolor=arcLink}
\fi
\printcontents[appendix]{}{1}[3]{}
\endgroup
\bigskip
\newpage

\section{\name-TRACE and \name-TRACE-DROID details}

\subsection{ARC-Trace: Data Labeling Details}
\label{app:labeling}

ARC-Trace is constructed through three stages, illustrated in Fig.~\ref{arch_labeling}. Gemini 2.5 Pro interprets the episode video and selects important frames. GPT-5.5 then generates episode-specific state checks and combines this evidence with the task, episode narrative, and selected observations to produce action-grounded causal traces. Separating episode-level interpretation from frame-level annotation gives each trace the context needed to explain the current decision.

\textbf{Source data and coverage.}
We use the training split of DROID v1.0.1 \citep{droid}, containing 95,658 episodes in RLDS/TFDS format. Demonstrations are recorded at 15,Hz and include wrist and two exterior camera views, robot states, actions, and task instructions. The labeling run covers approximately 74,740 episodes, or 78.1\% of the source split, and produces approximately 1.2 million annotated frames. This corresponds to roughly 16.1 annotations per labeled episode. Most episodes without labels lack a usable task instruction; processing failures are recorded separately.

\textbf{Episode narration and keyframe selection.}
Gemini 2.5 Pro receives the task instruction and episode video and produces a temporally grounded narrative of the demonstration. The narrative describes the objects involved, the robot's actions, and the resulting changes in the scene. This episode-level account provides context that an isolated image cannot establish, including which parts of the task have already been completed.
The same video-labeling stage selects important frames for annotation. Selection is guided by the VLM's interpretation of the demonstration, rather than a fixed uniform sampling interval. Each selected frame is mapped back to its original frame index $f$ and timestamp $f/15$.

\textbf{Episode-specific predicate generation.}
For the selected observations, GPT-5.5 generates state checks describing the objects, relations, and conditions relevant to the task. We use the low reasoning-effort setting for this stage. In the example in Fig.~\ref{arch_labeling}, these checks include \texttt{target\_object}, \texttt{target\_container}, \texttt{pot\_visible}, \texttt{gripper\_open} They identify the object and destination named by the instruction and make relevant visual conditions explicit before the final trace is composed.
Their outputs provide structured evidence to the trace labeler, reducing the need to reconstruct the task state from the image and narrative alone. These are model-generated descriptions of visual evidence, not independently measured ground-truth states.

\textbf{Action-grounded trace generation.}
GPT-5.5, again using low reasoning effort, combines the task instruction, episode narrative, selected wrist observation, and state-check outputs into a structured reasoning annotation. The annotation contains the seven semantic fields defined in Sec.~\ref{sec:reasoning_representation}, linking the perceived state and relevant cause to a possible consequence, chosen effect, action implication, forbidden effect, and completion status.
Each record includes a \texttt{natural\_language\_construction}, a fluent paragraph of three to six sentences targeting approximately 700 characters. This paragraph combines the causal account with the relevant object and grasp information. The structured fields support inspection and representation ablations, while the paragraph provides the natural-language reasoning used for policy conditioning.

\textbf{Temporal alignment and saved artifacts.}
Each annotation retains its episode index, keyframe identifier, original frame index, and timestamp. The episode index identifies the source record in the RLDS training split, and the frame index identifies the corresponding observation and start of the associated action chunk. This preserves alignment with the original demonstration without modifying its observations or actions.
We store one JSON record per episode, with annotations ordered chronologically. Selected images, episode videos, raw model responses, parsed outputs, model identifiers, and prompt versions are retained alongside the labels. The full episode video is used offline to construct supervision; it is not an additional input to the deployed policy.

\begin{figure}[p]
    \centering
    \includegraphics[
        width=\linewidth,
        height=0.90\textheight,
        keepaspectratio
    ]{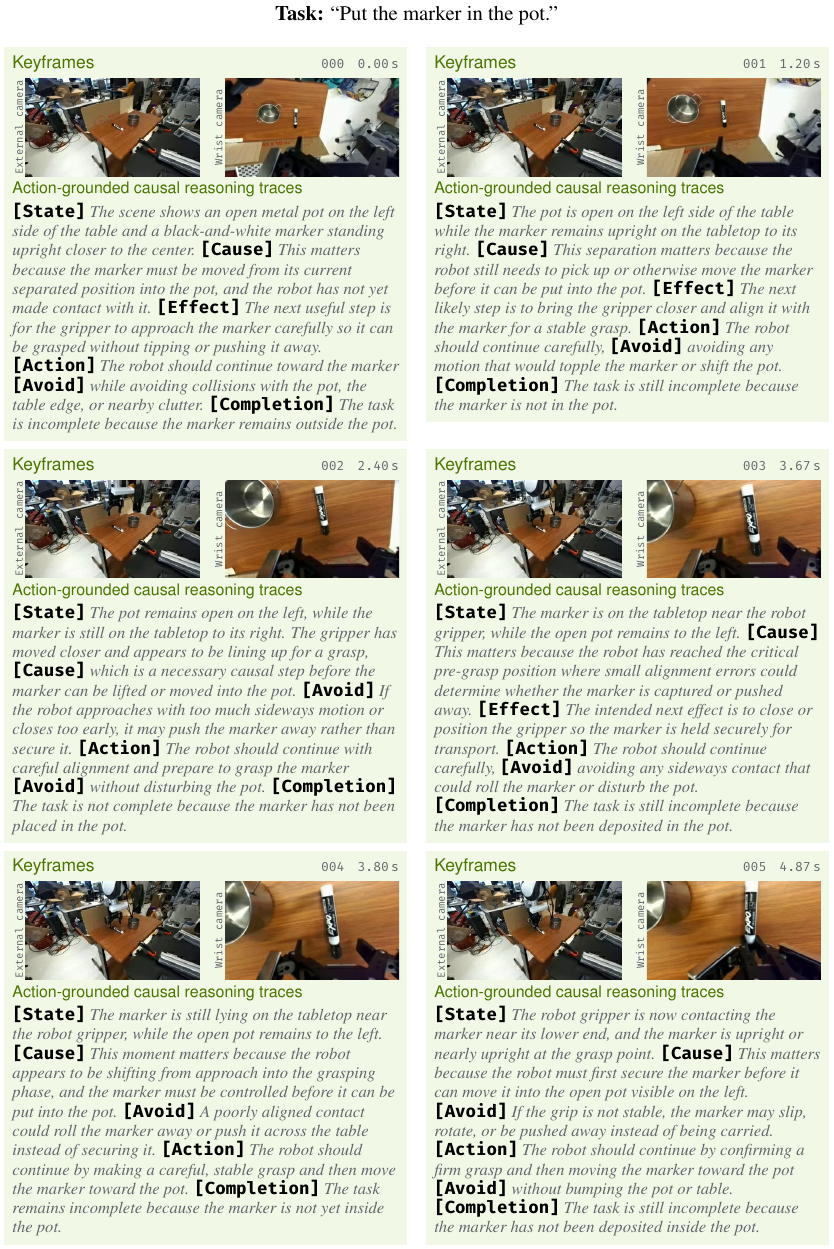}
    \caption{Full natural-language reasoning annotations for
    ``Put the marker in the pot,'' with synchronized external-
    and wrist-camera views. Part 1: labeled keyframes 000--005.}
    \label{fig:reasoning-trace-part1}
\end{figure}

\begin{figure}[p]
    \centering
    \includegraphics[
        width=\linewidth,
        height=0.90\textheight,
        keepaspectratio
    ]{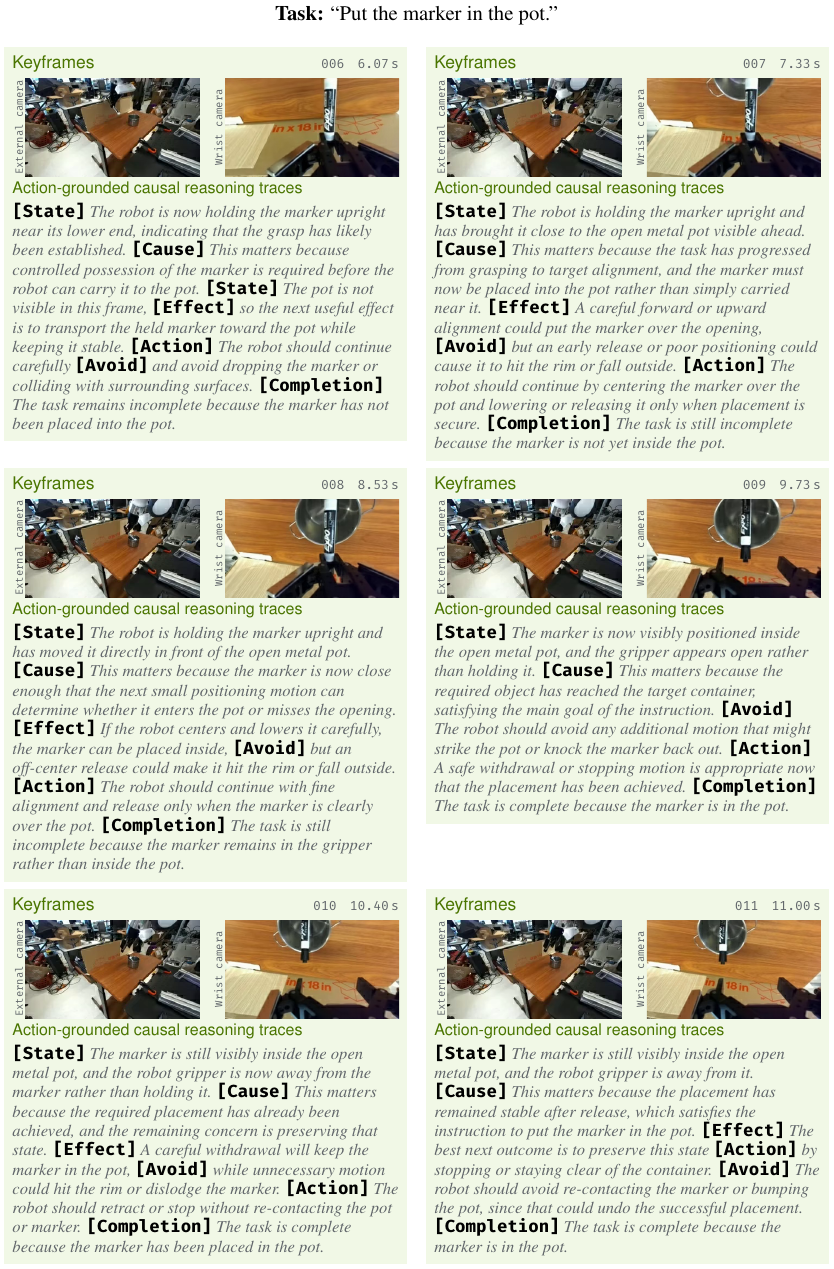}
    \caption{Full natural-language reasoning annotations for
    ``Put the marker in the pot,'' continued.
    Part 2: labeled keyframes 006--011.}
    \label{fig:reasoning-trace-part2}
\end{figure}

We show an annotated example from a randomly selected DROID episode in Figures \ref{fig:reasoning-trace-part1} and \ref{fig:reasoning-trace-part2}.

\clearpage

\clearpage

\section{Fine-tuning and inference recipe details}

\subsection{VLA Fine-Tuning Details}
\label{app:finetuning}
\begin{figure}[H]
    \centering
    \includegraphics[width=0.7\linewidth]{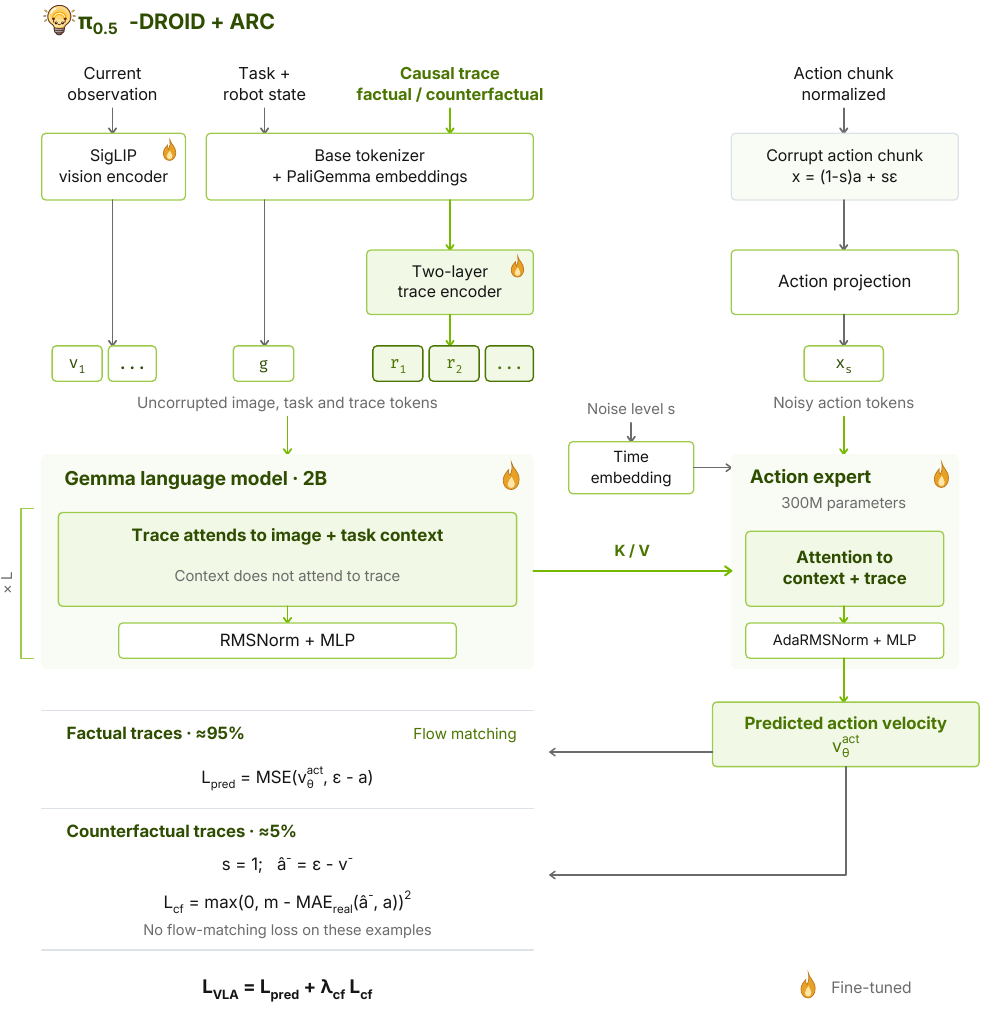}
    \caption{\name fine-tuning architecture for $\pi_{0.5}$}
    \label{fig:vla_architecture}
\end{figure}

\paragraph{Observation and instruction encoding.}
Figure~\ref{fig:vla_architecture} shows the $\pi_{0.5}$-DROID implementation. The exterior and wrist images in $o_t$ are resized to $224\times224$ and encoded by SigLIP into 256 tokens per view. These tokens are projected to the Gemma backbone's width of 2048. The task instruction $g$ and discretized robot state are encoded together using the PaliGemma tokenizer, with a maximum length of 200 tokens. Image and instruction tokens form the original context; unused camera slots and padding are masked.

\paragraph{Trace encoding.}
The trace $r_t$ is tokenized separately, with a maximum length of 320 tokens, and embedded using the same PaliGemma embedding table. The trace encoder projects these 2048-dimensional embeddings to width 512, adds learned positional embeddings, and applies two pre-normalized transformer blocks, each with eight attention heads and a feed-forward width of 2048. A final projection returns the representations to width 2048 before they are appended to the context. This projection is zero-initialized, so the added representations initially carry no trace-specific information.

\paragraph{Layerwise conditioning.}
Context and trace tokens are processed by the Gemma-2B parameters, while noisy action tokens are processed by the 300M-parameter action expert. The two components have widths 2048 and 1024, respectively, and interact through shared attention at each of the 18 transformer layers. Their attention projections produce compatible keys, queries, and values despite the different hidden widths. The block mask is

\begin{center}
  \includegraphics[width=0.28\linewidth]{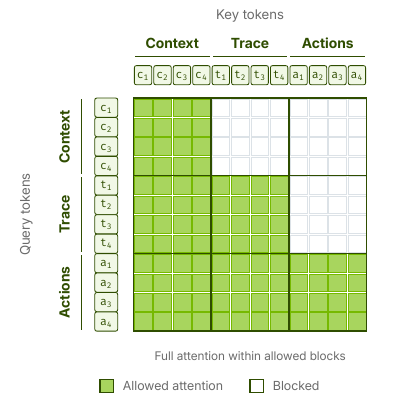}
\end{center}

where rows denote the attending tokens, columns denote the attended tokens, and $1$ permits attention to all valid tokens in the corresponding block. Thus, the trace can use the observed scene and instruction, while the action expert can combine both with the noisy action chunk. Neither context nor trace representations depend on the noisy actions. Although trace tokens precede action tokens in the input sequence, their rotary position indices are assigned after the action positions, preserving the pretrained positional indices of the action tokens.

\paragraph{Action prediction and training.}
With chunk length $H=15$, the normalized action target is $\mathbf{y}_t^{\mathrm{act}}=\mathbf{a}_t\in\mathbb{R}^{H\times32}$. Eight coordinates represent the seven joints and gripper; the remaining coordinates are padding. Each row of $\mathbf{x}_t^{\mathrm{act}}$ is projected to one 1024-dimensional action token. A sinusoidal embedding of $s_{\mathrm{act}}$, followed by an MLP, conditions the action expert through adaptive RMS normalization. Its final token representations are projected back to 32 coordinates to produce $v_{\theta}^{\mathrm{act}}$.

The vision encoder, language backbone, trace encoder, and action expert are jointly optimized with Eq.~\ref{eq:glaux_vla}. Factual and counterfactual examples share one batched forward pass. Their loss masks select flow matching or the counterfactual penalty, respectively, with $s_{\mathrm{act}}=1$ for counterfactual examples. The flow loss uses the padded action representation, whereas $\operatorname{MAE}_{\mathrm{real}}$ excludes padding when measuring counterfactual separation.

\subsection{WAM Fine-Tuning Details}
\label{app:cosmos}
\begin{figure}[t]
    \centering
    \includegraphics[width=0.7\linewidth]{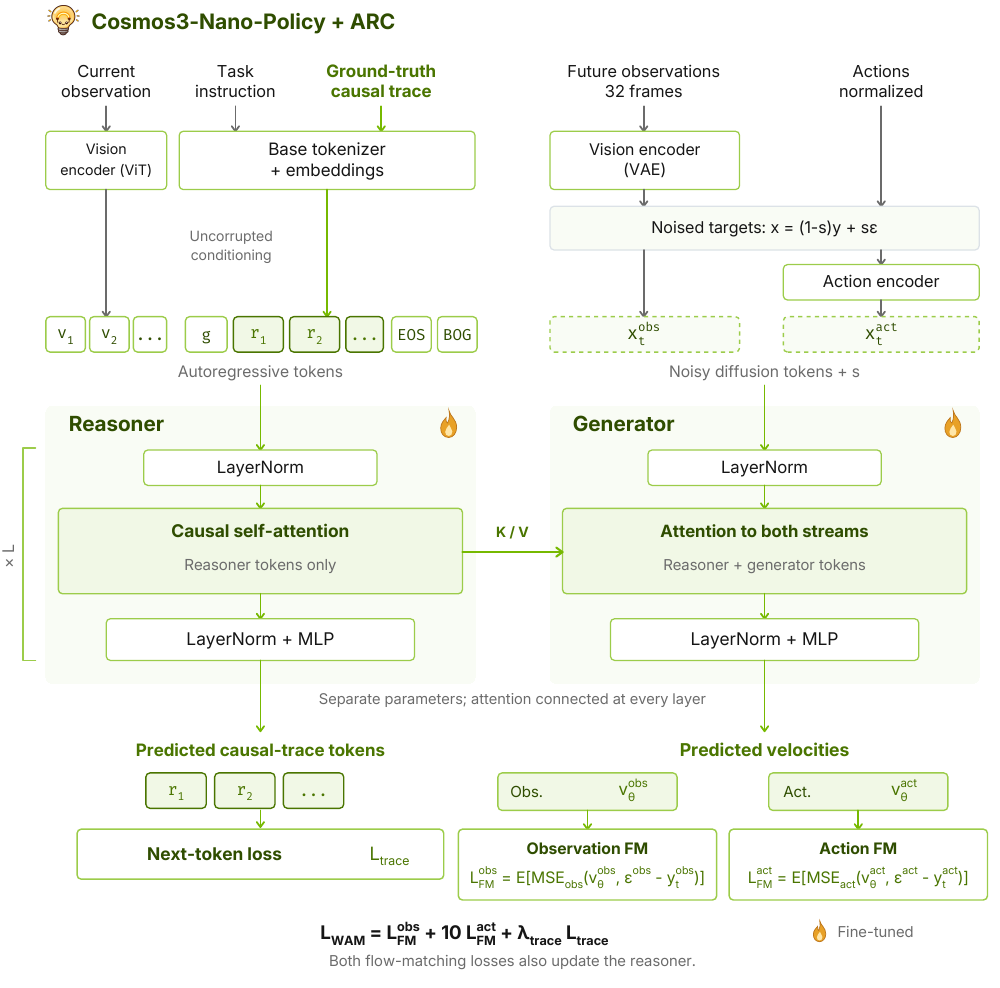}
    \caption{\name fine-tuning architecture for Cosmos3-nano-policy}
    \label{fig:wam_architecture}
\end{figure}

\paragraph{Token construction.}
Figure~\ref{fig:wam_architecture} shows the Cosmos3-Nano-Policy implementation. The reasoner receives ViT-encoded current observations followed by the task instruction $g$ and trace $r_t$, using the pretrained language tokenizer and embeddings. The trace is appended directly to the instruction without a separate trace encoder. End-of-sequence and begin-of-generation tokens mark the boundary between the reasoner and generator sequences.

The generator operates on continuous observation and action representations. Each training window contains one observed frame and 32 future frames with aligned actions. The video VAE encodes the future observations into $\mathbf{y}_t^{\mathrm{obs}}$, while the normalized actions form $\mathbf{y}_t^{\mathrm{act}}$. Their noisy versions, $\mathbf{x}_t^{\mathrm{obs}}$ and $\mathbf{x}_t^{\mathrm{act}}$, are projected into generator tokens. The base Cosmos video input also retains the initial frame as clean VAE conditioning, separate from the noised future targets \citep{cosmos3}.

\paragraph{Reasoner--generator attention.}
Cosmos3-Nano uses 36 paired transformer layers of width 4096. The reasoner and generator have separate normalization layers, attention projections, and feed-forward networks, but their tokens interact through a shared attention operation \citep{cosmos3}. The block mask is
\begin{center}
  \includegraphics[width=0.28\linewidth]{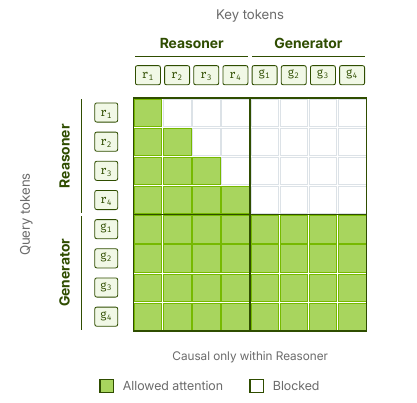}
\end{center}
using the same row and column convention as the VLA. Causal attention restricts each reasoner token to its preceding context and itself. Generator tokens attend to the complete reasoner sequence and to one another, allowing action and future-observation predictions to exchange information while remaining conditioned on the trace. Separate output projections produce $v_{\theta}^{\mathrm{act}}$ and $v_{\theta}^{\mathrm{obs}}$ in their respective target spaces.

\paragraph{Joint supervision.}
The ground-truth trace serves two roles during training. It supplies conditioning to the generator and next-token targets to the reasoner. For $\mathcal{L}_{\mathrm{trace}}$, targets are shifted so that prediction of $r_{t,j}$ uses only $o_t$, $g$, and $r_{t,<j}$. The reasoner cannot attend to later trace tokens or diffusion targets. The generator, however, conditions on the complete ground-truth trace when predicting actions and future observations.

We jointly update both components using Eq.~\ref{eq:glaux_wam}. The trace loss supervises the reasoner's language predictions, while $\mathcal{L}_{\mathrm{pred}}$ also updates its representations through the keys and values used by the generator. Consequently, the reasoner is trained both to express the annotated reasoning and to provide features useful for continuous prediction. The counterfactual objective is omitted, as described in Sec.~\ref{sec:wam_finetuning}.

\subsection{Inference}
\label{app:external_vlm}

During evaluation, we freeze the ARC-fine-tuned policies and use an external VLM to generate action-grounded causal traces. Each request contains the task instruction and the current exterior- and wrist-camera images. A fixed prompt specifies the ARC trace format and requests a single natural-language paragraph of at most 900 characters, grounded in the current observations.

The generated trace is supplied alongside the original instruction and observations. In the VLA, the vision-language backbone encodes the trace for the action expert; in the WAM, the reasoner encodes it for the generator. The external VLM provides the reasoning, while the ARC-fine-tuned policy remains responsible for generating continuous robot actions. Responses are checked for format and length before use, and a rejected response leaves the previous accepted trace unchanged. Trace generation runs asynchronously, with each policy call using the latest accepted trace. 

\subsubsection{Real-Time Deployment}
\label{app:inference}
\begin{figure}[H]
    \centering
    \includegraphics[width=1\linewidth]{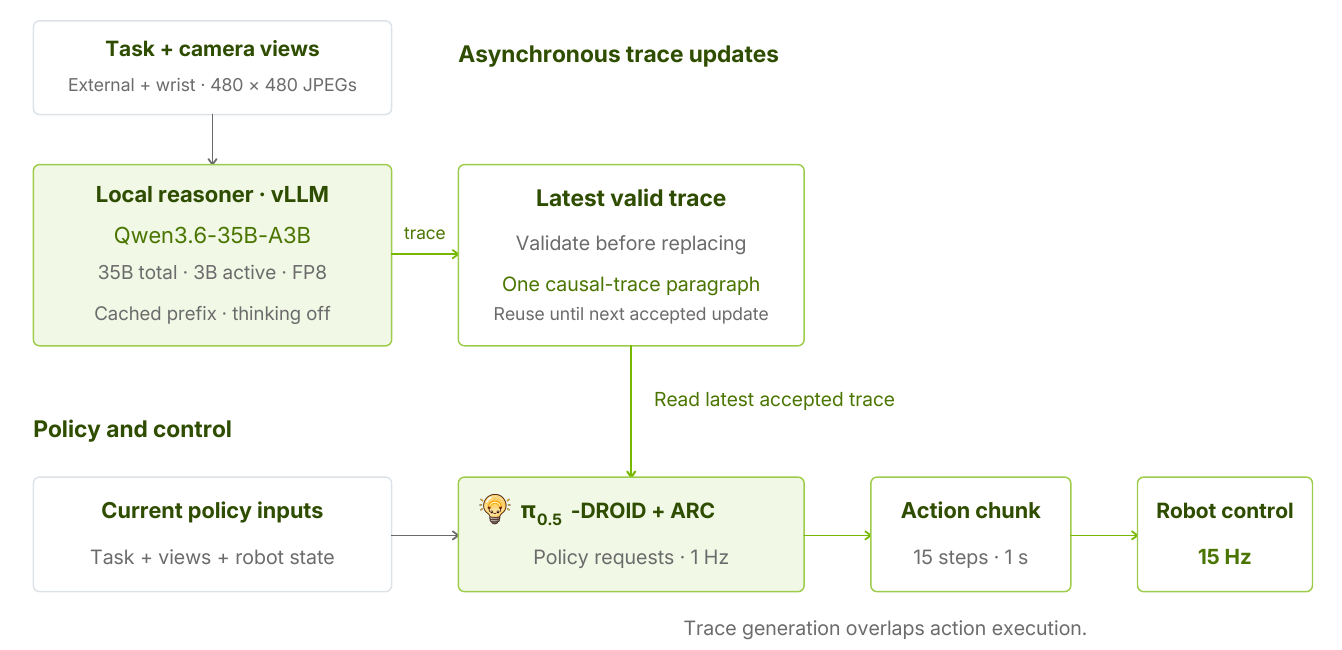}
    \caption{Real-time deployment strategy}
    \label{fig:placeholder}
\end{figure}
We separate reasoning generation from action execution so that the robot does not wait for a new trace at every policy query. The robot executes 15-action chunks at 15\,Hz, while a background thread generates reasoning from the latest camera observations. After episode initialization, each policy query uses the most recent validated trace; delayed or invalid responses leave the previous trace in place. Reasoning latency therefore affects how recently the scene was interpreted, without blocking subsequent action generation.

To refresh the reasoning within the action-chunk duration, we serve Qwen3.6-35B-A3B-FP8 locally using vLLM on a dedicated GPU host. Local serving avoids remote API delays, while FP8 inference and prefix caching reduce computation and memory costs. The shared system prompt and task instruction precede the images, allowing their cached representations to be reused across requests. We disable auxiliary thinking-token generation and request a single causal paragraph of at most 900 characters. Each request contains two camera images, with bounded concurrency to limit queueing.

\section{Experiment Details}

\subsection{Simulation Benchmark Details}

\subsubsection{RoboLab-120}

We evaluate ARC-fine-tuned $\pi_{0.5}$ and Cosmos3-Nano-Policy on all 120 tasks in RoboLab-120 \citep{robolab}, using the released benchmark without modifying its tasks, scenes, or success criteria. Evaluation uses the DROID embodiment and follows the standard protocol of 10 episodes per task. We repeat this evaluation for the vague, default, and specific instruction variants, yielding 1,200 episodes per model for each instruction setting. We report success as the fraction of episodes that satisfy the benchmark's task-completion criteria. All evaluations are zero-shot, with model weights frozen and no training on benchmark data.
We refer the readers to the original paper \cite{robolab} for the full dataset split and examples.

\subsubsection{MolmoSpaces}

We evaluate the same checkpoints on all nine manipulation tasks reported in Table~\ref{tab:arc_molmospaces}, following the official MolmoSpaces evaluation protocol \citep{kim2026molmospaces}. We use the released episode sets and task configurations, preserving the specified scenes, object placements, camera configurations, instructions, and success criteria. All tasks use the DROID embodiment, with no benchmark-specific training or modification of the evaluation environments. We report success for each task and the aggregate performance using the benchmark's scoring procedure.
We refer the readers to the original paper \cite{kim2026molmospaces} for the full dataset split and examples.

For both benchmarks, results for the other baseline policies are taken from the corresponding benchmark reports and official leaderboards.

\subsubsection{RoboLab-Reasoning-50}
\label{app:robolab_reasoning50}

We build RoboLab-Reasoning-50 using the RoboLab framework, introducing 50 tasks that require target inference, multi-step execution, discovery, or reasoning about exclusions. Of these, 34 leave the target unnamed and require the policy to identify it from the instruction and scene. The remaining 16 name their targets but require long-horizon or discovery behavior.

The benchmark combines 27 tasks adapted from our real-world evaluation with 23 simulation-only tasks that extend the object sets and destinations. The paired tasks preserve the instruction wording used on hardware. Tasks span 44 scene and use objects from a processed library of 371 NVIDIA Isaac objects. The robot and camera configuration match the DROID embodiment used in RoboLab-120.

\textbf{Evaluation.}
We run 10 episodes for each of the 50 tasks, totaling 500 episodes per
model. Control runs at 15\,Hz, with task-specific time limits ranging
from 50 to 240\,s. Budgets allow at least 45\,s for the first required
move and 30\,s for each additional move, with a further 30\,s for
discovery tasks. An episode terminates on success, timeout, or a
tracked object leaving the workspace.

\textbf{Scoring.}
Headline results use binary task success, evaluated from the final
scene state with the gripper detached. Predicates check containment,
placement, stacking, exclusion constraints, or task-specific
combinations of these conditions. Scoring does not impose a fixed
execution sequence, although the required final arrangement must be
satisfied. 

\begingroup
\definecolor{RFTGreen}{HTML}{76B900}
\definecolor{RFTDeep}{HTML}{2F4D00}
\definecolor{RFTTint}{HTML}{F1F8E6}
\definecolor{RFTAccent}{HTML}{A8D55E}
\definecolor{RFTGray}{HTML}{6B6B6B}
\definecolor{RFTTaskGray}{HTML}{92989F}
\definecolor{RFTRule}{HTML}{DCE2E7}
\setlength{\tabcolsep}{0pt}
\setlength{\arrayrulewidth}{0.3pt}
\setlength{\LTleft}{0pt}\setlength{\LTright}{0pt}
\setlength{\LTcapwidth}{\linewidth}
\renewcommand{\arraystretch}{1}
\def\RFTFont{\fontencoding{T1}\fontfamily{FiraMono-TLF}\fontseries{m}\fontshape{n}\fontsize{9}{11}\selectfont}
\long\def\RFTTask#1{%
  \fontencoding{T1}\fontfamily{phv}\fontseries{m}\fontshape{it}%
  \fontsize{9}{11}\selectfont
  \color{RFTTaskGray}\raggedright\arraybackslash
  \textquotedblleft#1\textquotedblright\strut}
\def\RFTCategory#1{\RFTFont\fontsize{8.5}{11}\selectfont\color{RFTGray}\raggedright\arraybackslash #1\strut}
\def\RFTHeader#1{{\RFTFont\bfseries\fontsize{9}{11}\selectfont\color{RFTDeep}#1}}
\def\RFTTitle#1{{\RFTFont\bfseries\fontsize{10}{12}\selectfont\color{RFTDeep}#1}}
\begin{longtable}{@{}p{\dimexpr\linewidth-88pt\relax}@{\hspace{8pt}}p{80pt}@{}}
\caption{RoboLab-Reasoning-50 tasks and categories. Tasks assigned to two categories retain both labels.}\label{tab:reasoning50-tasks}\\
\rowcolor{RFTTint}\multicolumn{2}{@{}p{\linewidth}@{}}{\RFTTitle{RoboLab-Reasoning-50}\hfill{\RFTFont\textcolor{RFTGray}{50 tasks}}}\\[5pt]
\RFTHeader{Task}&\RFTHeader{Category}\\[3pt]
\arrayrulecolor{RFTAccent}\hline\arrayrulecolor{RFTRule}
\endfirsthead
\rowcolor{RFTTint}\multicolumn{2}{@{}p{\linewidth}@{}}{\RFTTitle{RoboLab-Reasoning-50}\hfill{\RFTFont\textcolor{RFTGray}{continued}}}\\[5pt]
\RFTHeader{Task}&\RFTHeader{Category}\\[3pt]
\arrayrulecolor{RFTAccent}\hline\arrayrulecolor{RFTRule}
\endhead
\multicolumn{2}{r}{{\RFTFont\fontsize{6.5}{8}\selectfont\textcolor{RFTGray}{Continued on next page}}}\\
\endfoot
\endlastfoot
\RFTTask{The plate should only have edible items} & \RFTCategory{Simple} \\[3pt]\hline
\RFTTask{Everything made of wood goes in the bowl} & \RFTCategory{Simple} \\[3pt]\hline
\RFTTask{Put the smallest item in the bowl} & \RFTCategory{Simple} \\[3pt]\hline
\RFTTask{Put the second tallest object in the bowl} & \RFTCategory{Simple} \\[3pt]\hline
\RFTTask{Put the item that matches the bowl's color into it} & \RFTCategory{Simple} \\[3pt]\hline
\RFTTask{Move the object closest to the table's edge to the bowl or plate} & \RFTCategory{Simple} \\[3pt]\hline
\RFTTask{Put the kitchen utensils in the plate} & \RFTCategory{Simple} \\[3pt]\hline
\RFTTask{Stack the blocks in order: red, blue, green} & \RFTCategory{Long horizon} \\[3pt]\hline
\RFTTask{Reverse the stacking order of the three colored blocks} & \RFTCategory{Long horizon} \\[3pt]\hline
\RFTTask{Reverse the arrangement: whatever is on the plate goes in the bowl, and whatever is in the bowl goes on the plate} & \RFTCategory{Long horizon} \\[3pt]\hline
\RFTTask{The plate is for utensils, the bowl is for blocks; sort the table accordingly} & \RFTCategory{Long horizon} \\[3pt]\hline
\RFTTask{Nothing should be left on the left plate} & \RFTCategory{Long horizon} \\[3pt]\hline
\RFTTask{Every container on the table should hold exactly two things} & \RFTCategory{Long horizon} \\[3pt]\hline
\RFTTask{Put the green vegetables and a fruit in the blue basket} & \RFTCategory{Long horizon} \\[3pt]\hline
\RFTTask{Put the drink cans in the cardboard box} & \RFTCategory{Long horizon} \\[3pt]\hline
\RFTTask{Pack all the bread into the basket} & \RFTCategory{Long horizon} \\[3pt]\hline
\RFTTask{Put the apples in the cardboard box} & \RFTCategory{Long horizon} \\[3pt]\hline
\RFTTask{Put all the citrus fruits into the basket} & \RFTCategory{Long horizon} \\[3pt]\hline
\RFTTask{Put the cups onto the tray, upright} & \RFTCategory{Long horizon} \\[3pt]\hline
\RFTTask{Put the yellow things onto the yellow plate and the green things onto the green plate} & \RFTCategory{Long horizon} \\[3pt]\hline
\RFTTask{Put all the fruits except the red ones on the higher plate} & \RFTCategory{Long horizon\newline + Negation} \\[3pt]\hline
\RFTTask{Put two bags of chips in the green basket} & \RFTCategory{Long horizon} \\[3pt]\hline
\RFTTask{Empty the right half of the desk into the grey bin} & \RFTCategory{Long horizon} \\[3pt]\hline
\RFTTask{Put all the tapes into the bin, except the tape in the tape dispenser} & \RFTCategory{Long horizon\newline + Negation} \\[3pt]\hline
\RFTTask{Put the reusable cups in the bussing bin and put the disposable cups in the trash bin} & \RFTCategory{Long horizon} \\[3pt]\hline
\RFTTask{Put the cutlery in the bussing bin} & \RFTCategory{Long horizon} \\[3pt]\hline
\RFTTask{Put a zucchini, the eggplant, and a knife on the cutting board} & \RFTCategory{Long horizon} \\[3pt]\hline
\RFTTask{Sort everything on the table into the recycling, compost, and trash bins} & \RFTCategory{Long horizon} \\[3pt]\hline
\RFTTask{Sort the chips bag, the banana, the drink can, and the paper cup into the correct bins} & \RFTCategory{Long horizon} \\[3pt]\hline
\RFTTask{Put the food snacks in the black bin, and leave the drinks} & \RFTCategory{Long horizon} \\[3pt]\hline
\RFTTask{Put the number cubes in the basket} & \RFTCategory{Long horizon} \\[3pt]\hline
\RFTTask{Move the grey plate and put whatever is under it into the red bowl} & \RFTCategory{Discovery} \\[3pt]\hline
\RFTTask{Take the black plate off of the bowl and put what you find underneath onto the grey plate} & \RFTCategory{Discovery} \\[3pt]\hline
\RFTTask{Move the carton, then put whatever was hidden behind it into the bowl} & \RFTCategory{Discovery} \\[3pt]\hline
\RFTTask{Move the mustard bottle aside and put whatever was behind it into the bowl} & \RFTCategory{Discovery} \\[3pt]\hline
\RFTTask{Take the item out of the plate and place it on the table. Then put the item that was on the table onto the plate.} & \RFTCategory{Discovery} \\[3pt]\hline
\RFTTask{Put the banana on the plate} & \RFTCategory{Discovery} \\[3pt]\hline
\RFTTask{Put the spatula on the plate} & \RFTCategory{Discovery} \\[3pt]\hline
\RFTTask{Put the bananas on the plate} & \RFTCategory{Discovery} \\[3pt]\hline
\RFTTask{Put the cans in the recycling} & \RFTCategory{Discovery} \\[3pt]\hline
\RFTTask{Put what's under the plate into the grey bin} & \RFTCategory{Discovery} \\[3pt]\hline
\RFTTask{Stack the blue block behind the paddle on the rubik's cube} & \RFTCategory{Discovery} \\[3pt]\hline
\RFTTask{Pack the can behind the box into the box} & \RFTCategory{Discovery} \\[3pt]\hline
\RFTTask{One container is empty and one has something in it. Put the block into the empty container} & \RFTCategory{Negation} \\[3pt]\hline
\RFTTask{Put all the fruits into the bowl except the orange} & \RFTCategory{Negation} \\[3pt]\hline
\RFTTask{Put all items on the table into the bowl except the can} & \RFTCategory{Negation} \\[3pt]\hline
\RFTTask{Put every item on the table into the bowl except the blue block and the wood block} & \RFTCategory{Negation} \\[3pt]\hline
\RFTTask{Items are arranged in two groups on either side of the wall of blocks. Put all items from the group that does NOT contain a can into the bowl} & \RFTCategory{Negation} \\[3pt]\hline
\RFTTask{Three items are on the table: two are fruits and one is not. Put the non-fruit into the bowl} & \RFTCategory{Negation} \\[3pt]\hline
\RFTTask{Put every item on the table into the crate except the green colored objects} & \RFTCategory{Negation} \\[3pt]\hline
\end{longtable}
\endgroup

\subsection{RoboLab-Reasoning-Hardware}
\label{app:hardware}

\begin{figure}[H]
    \centering
    \includegraphics[width=0.4\linewidth]{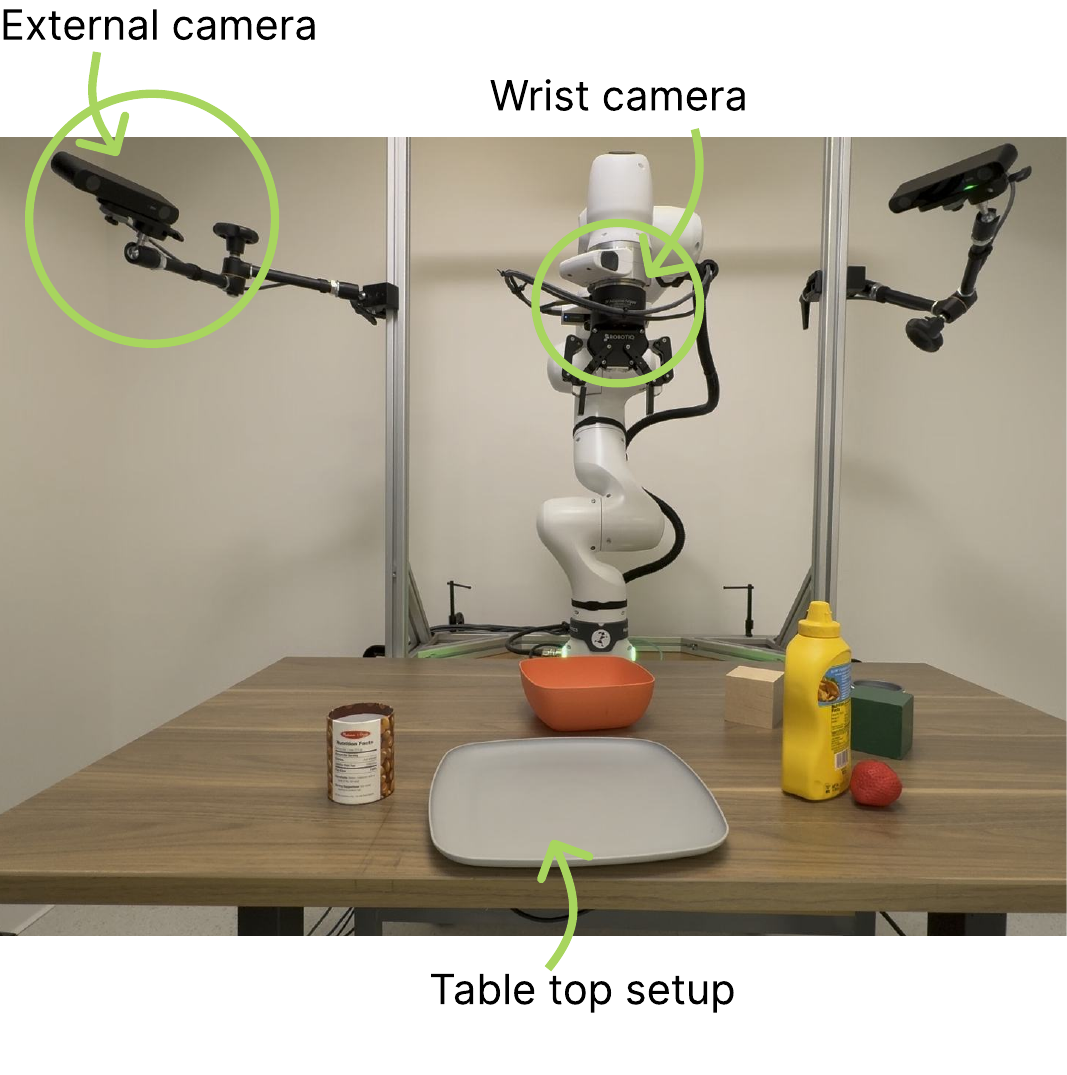}
    \caption{Hardware setup for real-world experiments}
    \label{fig:hardware_setup}
\end{figure}

We conduct hardware experiments on the DROID embodiment \citep{droid}, consisting of a seven-degree-of-freedom Franka Emika Panda arm equipped with a Robotiq 2F-85 two-finger gripper. Figure~\ref{fig:hardware_setup} shows the workspace layout and camera placements.

The evaluation includes simple reasoning, long-horizon, discovery, and negation tasks paired with their counterparts in RoboLab-Reasoning-50. Each pair uses the same task instruction, allowing direct comparison between simulation and physical execution. We evaluate ARC-fine-tuned $\pi_{0.5}$ and the comparison policies for three episodes per task, using the same checkpoints evaluated in simulation without hardware-specific fine-tuning. 

\begin{figure}[H]
    \centering
    \includegraphics[width=\linewidth,height=0.90\textheight,keepaspectratio]{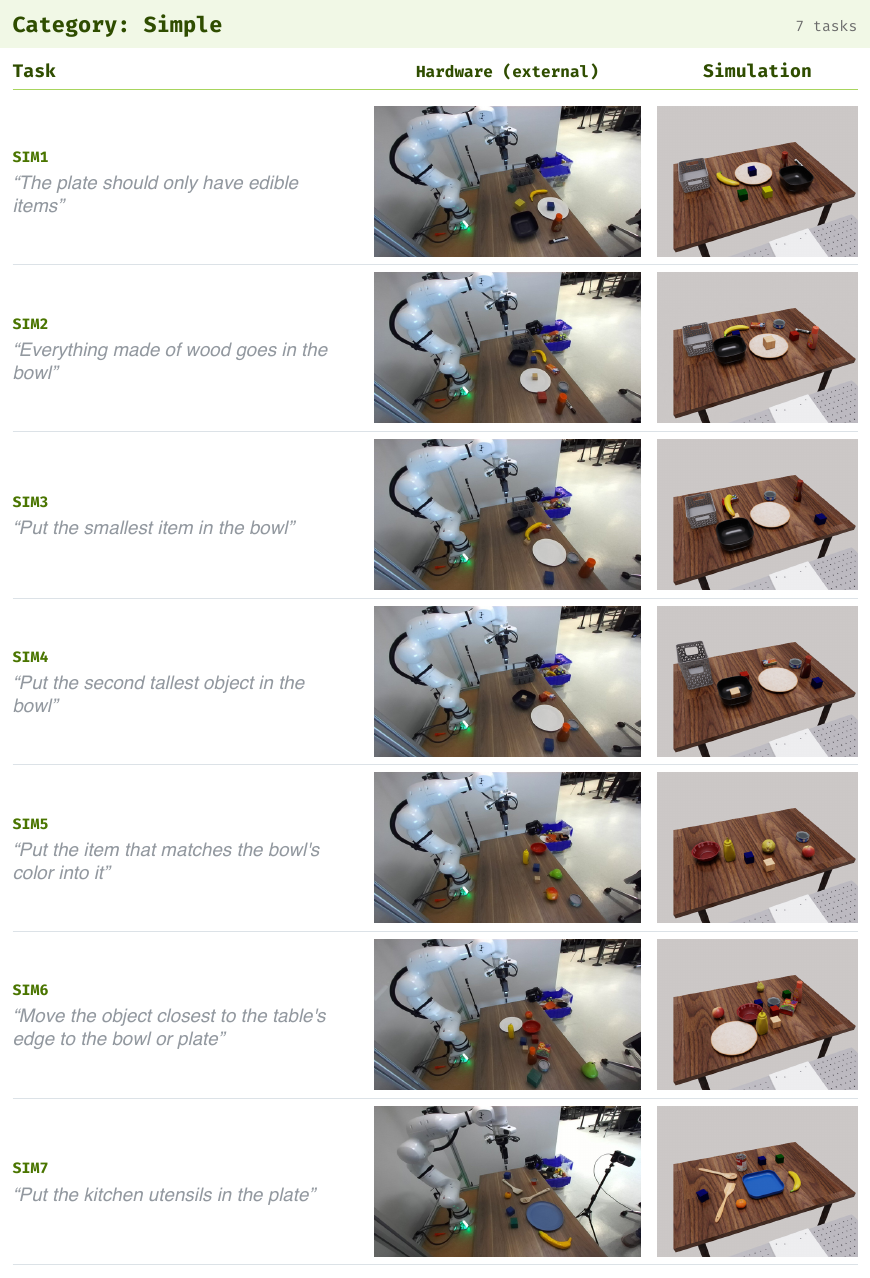}
    \caption{RoboLab-Reasoning-Hardware task setups for the simple category. Each row pairs the task instruction with the external hardware-camera view and its simulation setup.}
    \label{fig:hardware-tasks-simple}
\end{figure}

\begin{figure}[H]
    \centering
    \includegraphics[width=\linewidth,height=0.90\textheight,keepaspectratio]{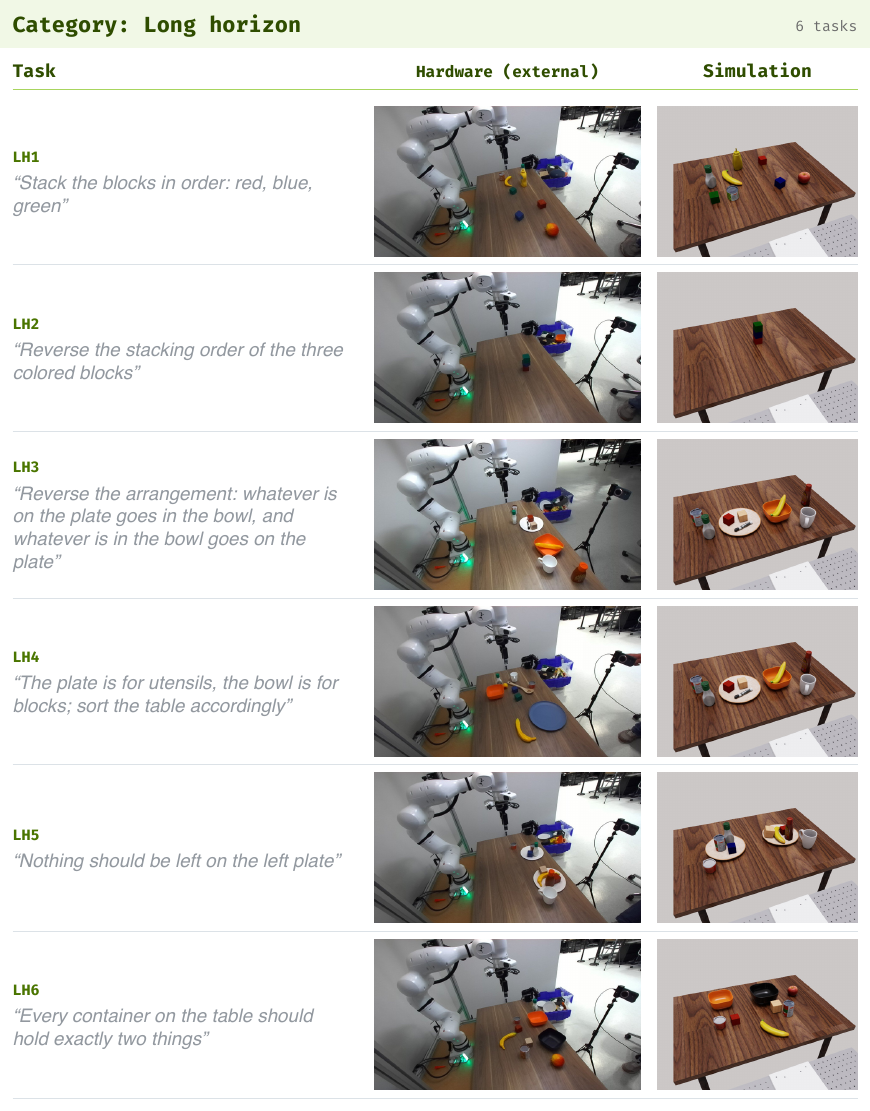}
    \caption{RoboLab-Reasoning-Hardware task setups for the long horizon category. Task instructions are shown alongside external hardware-camera views and simulation illustrations.}
    \label{fig:hardware-tasks-long-horizon}
\end{figure}

\begin{figure}[H]
    \centering
    \includegraphics[width=\linewidth,height=0.90\textheight,keepaspectratio]{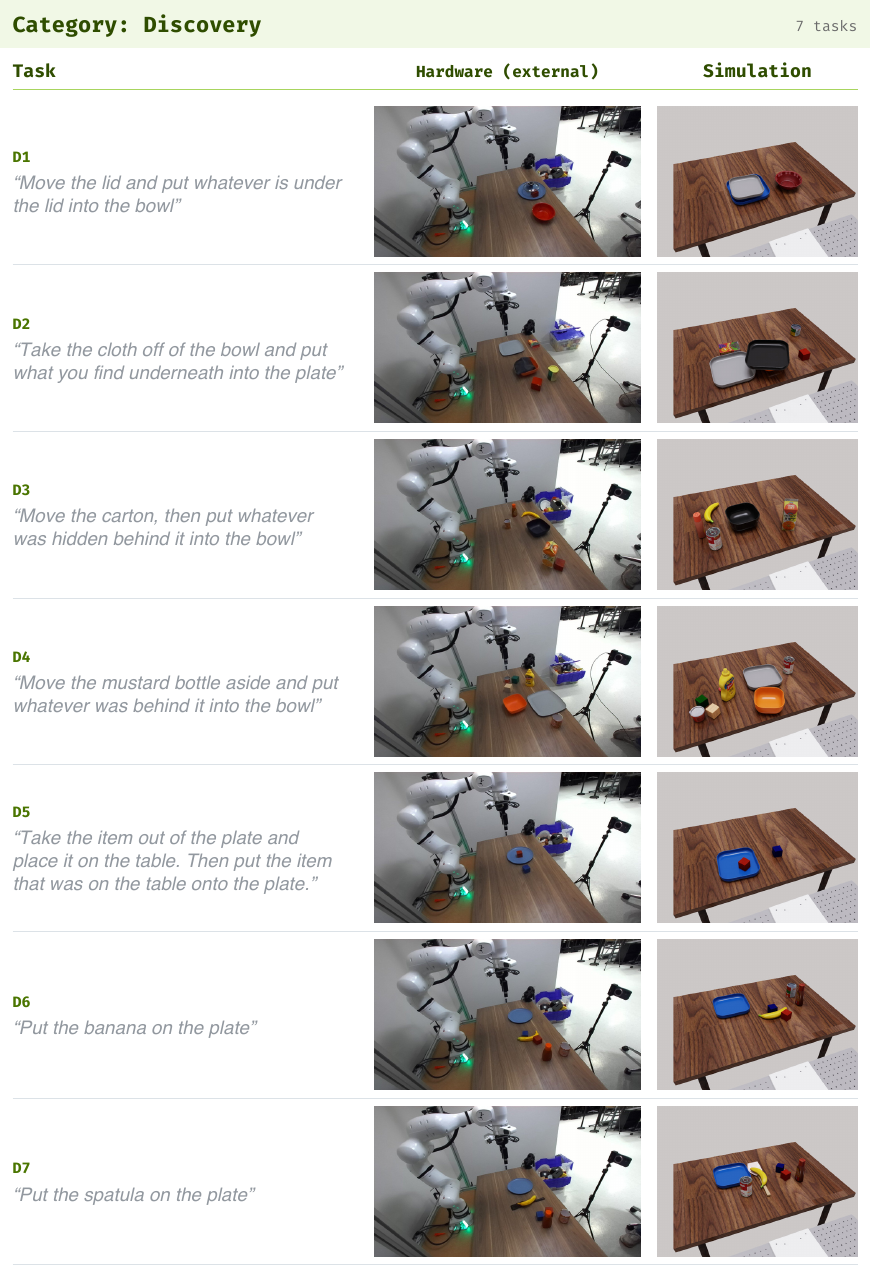}
    \caption{RoboLab-Reasoning-Hardware task setups for the discovery category. Each row pairs the task instruction with the external hardware-camera view and its simulation setup.}
    \label{fig:hardware-tasks-discovery}
\end{figure}

\begin{figure}[H]
    \centering
    \includegraphics[width=\linewidth,height=0.90\textheight,keepaspectratio]{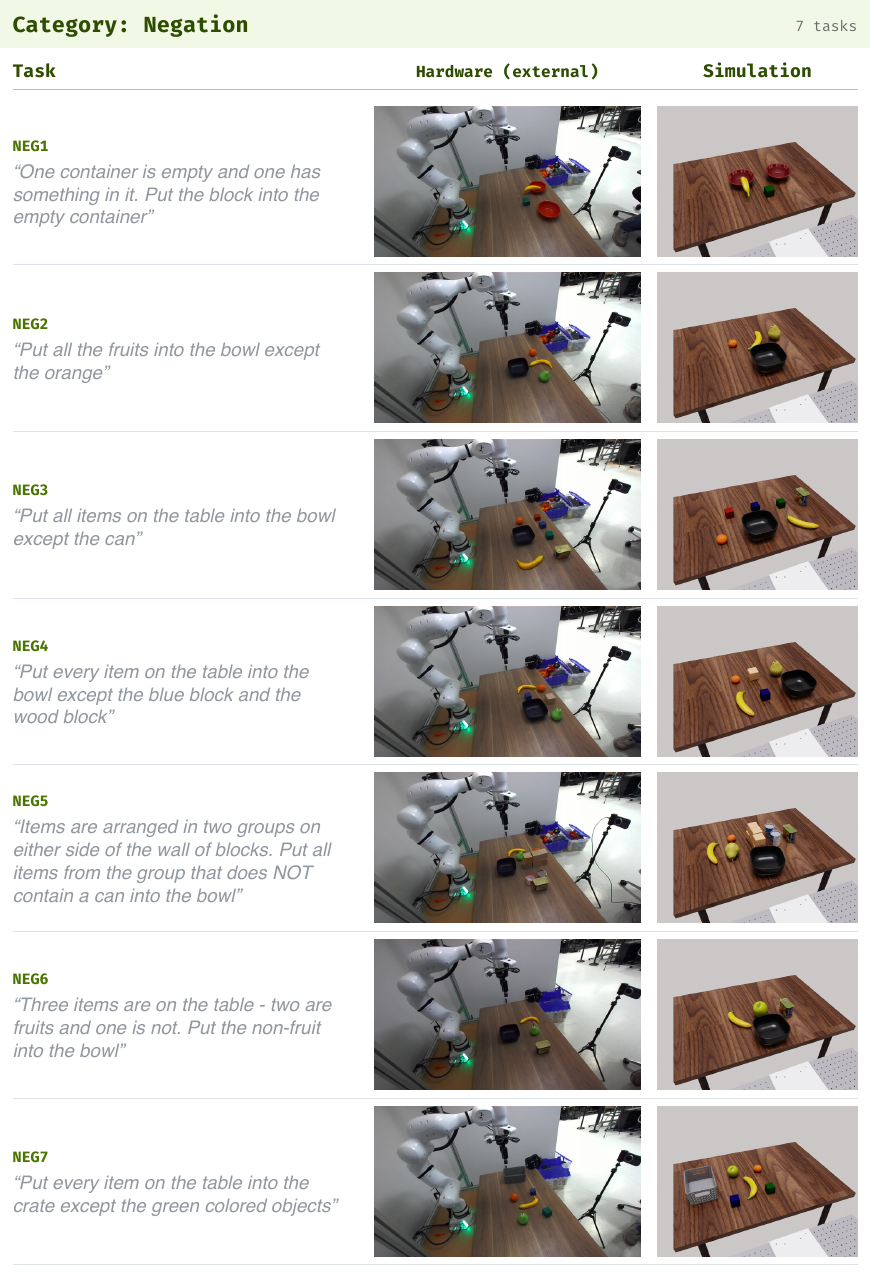}
    \caption{RoboLab-Reasoning-Hardware task setups for the negation category. Each row pairs the task instruction with the external hardware-camera view and its simulation setup.}
    \label{fig:hardware-tasks-negation}
\end{figure}

Figures \ref{fig:hardware-tasks-simple}, \ref{fig:hardware-tasks-long-horizon}, \ref{fig:hardware-tasks-discovery}, \ref{fig:hardware-tasks-negation} represents the benchmark tasks and the paired hardware and simulation environments. 

\clearpage

\clearpage

\clearpage

\clearpage

\subsection{Evaluated Policies}
\label{app:policies}

\paragraph{$\pi_{0.5}$.}
We use the released $\pi_{0.5}$-DROID joint-position checkpoint. Its PaliGemma backbone combines a SigLIP vision encoder with a Gemma language model to process camera observations, the task instruction, and tokenized robot state. A separate 300M-parameter action expert attends to these representations and predicts continuous action chunks through flow matching. The baseline receives the original task instruction without additional reasoning, using the default ten Euler steps for action generation.

\paragraph{Cosmos3-Nano-Policy.}
We use the DROID checkpoint of Cosmos3-Nano-Policy, a world-action model built on a mixture-of-transformers architecture. It combines a vision-language reasoner with a diffusion generator, each with separate parameters connected through attention at every layer. The reasoner processes the visual and language inputs, while the generator jointly predicts future observations and actions conditioned on the reasoner's representations. We refer readers to the original papers for detailed architecture and pretraining descriptions \citep{pi05,cosmos3}.

\paragraph{$\pi_{0.5}$ with explicit subtasks.}
This baseline evaluates external task decomposition without modifying or fine-tuning the policy. We retain the stock $\pi_{0.5}$-DROID checkpoint and use GPT-5.5 to generate an ordered plan from the task instruction and initial exterior-camera image. The plan contains at most 12 coarse subtasks, with a complete pick-and-place operation treated as one subtask. For multi-object instructions, the planner names the matching objects explicitly and excludes those ruled out by the instruction. Single-object and discovery instructions are passed through unchanged under this coarse decomposition scheme.

During execution, a client-side wrapper replaces the original instruction with the current subtask, leaving the observations, robot state, and policy unchanged. The client executes 15 actions per policy query. Every 30 simulation steps, GPT-5.5 checks the current subtask against the exterior and wrist-camera images. Completion requires the object to be at its destination, released, and stable; two consecutive positive checks advance the plan. A subtask also advances when its allocated step budget expires, with the episode budget divided across subtasks subject to a minimum of 225 steps per subtask. Failed subtasks are skipped rather than retried, and the final subtask remains active until the episode ends.

The planner and completion checker use low reasoning effort and a temperature of 0.2. The plan is generated once per episode, with no replanning; an empty or invalid plan falls back to the original instruction. The external VLM only supplies subtask prompts and completion judgments, while $\pi_{0.5}$ remains responsible for all robot actions.

\begingroup\setkeys{Gin}{height=0.25\textheight,keepaspectratio}\subsection{Detailed RoboLab-120 results.}
\label{detailed-app}
We now present detailed benchmark results under vague, default,
and specific instructions. Alongside success rates, we report
end-effector speed and EE SPARC, together with the available
uncertainty estimates. Figure~\ref{fig:robolab_detailed} compares
both ARC policies with the leading baselines.

\begin{figure}[H]
    \centering

    \includegraphics[width=\linewidth]
        {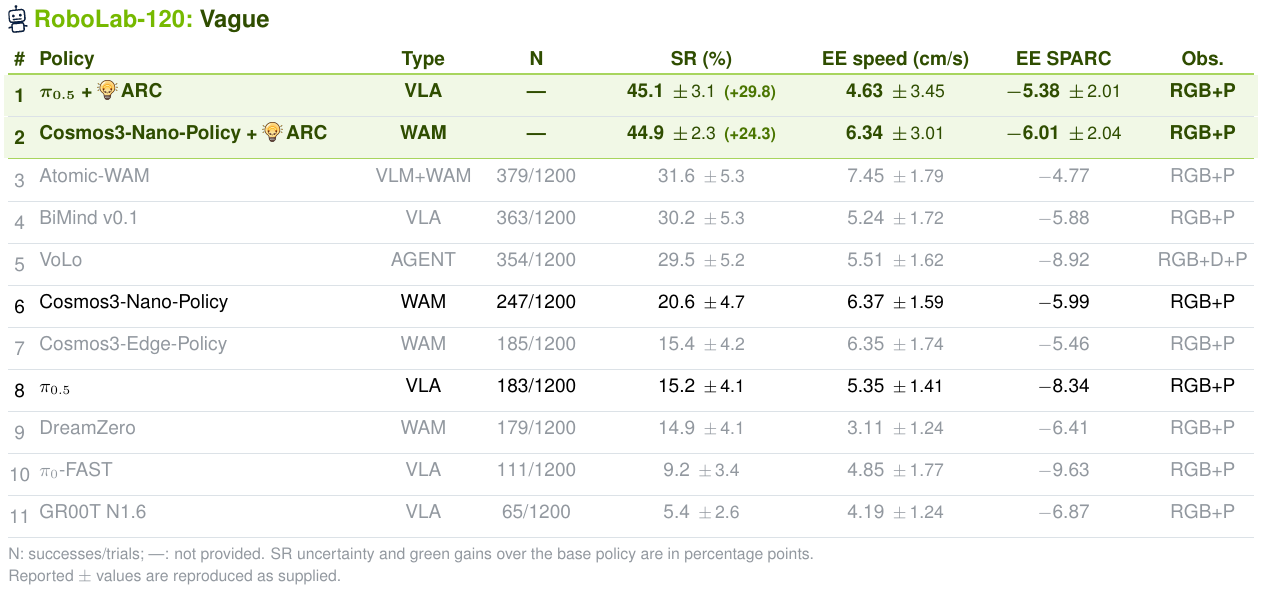}

    \vspace{6pt}

    \includegraphics[width=\linewidth]
        {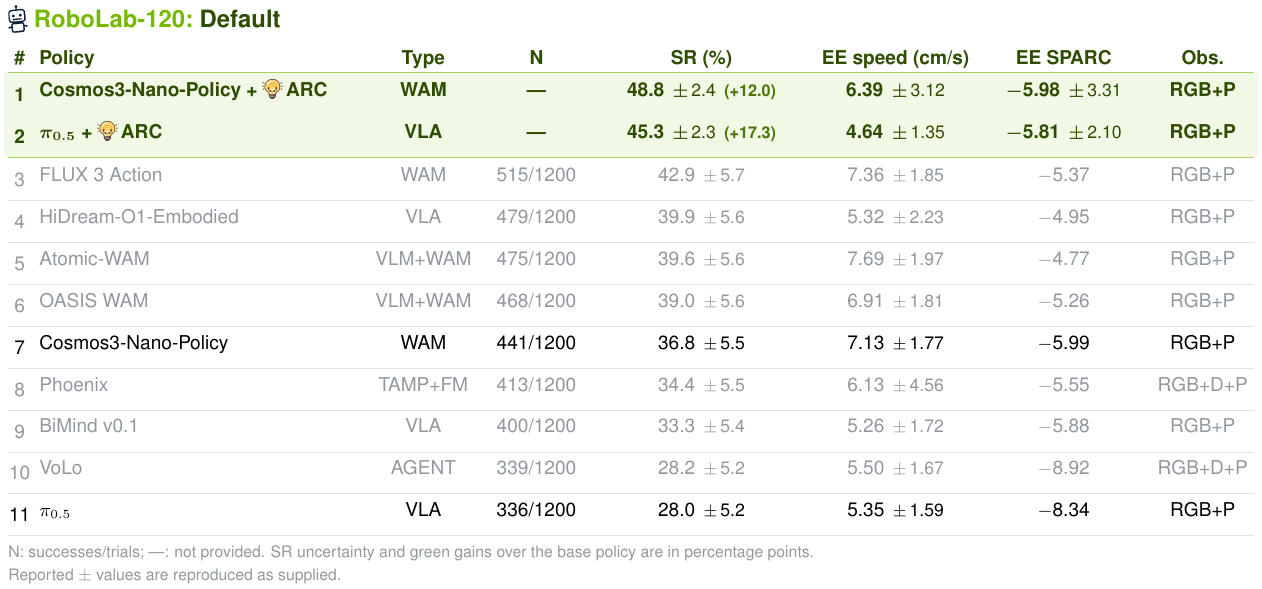}

    \vspace{6pt}

    \includegraphics[width=\linewidth]
        {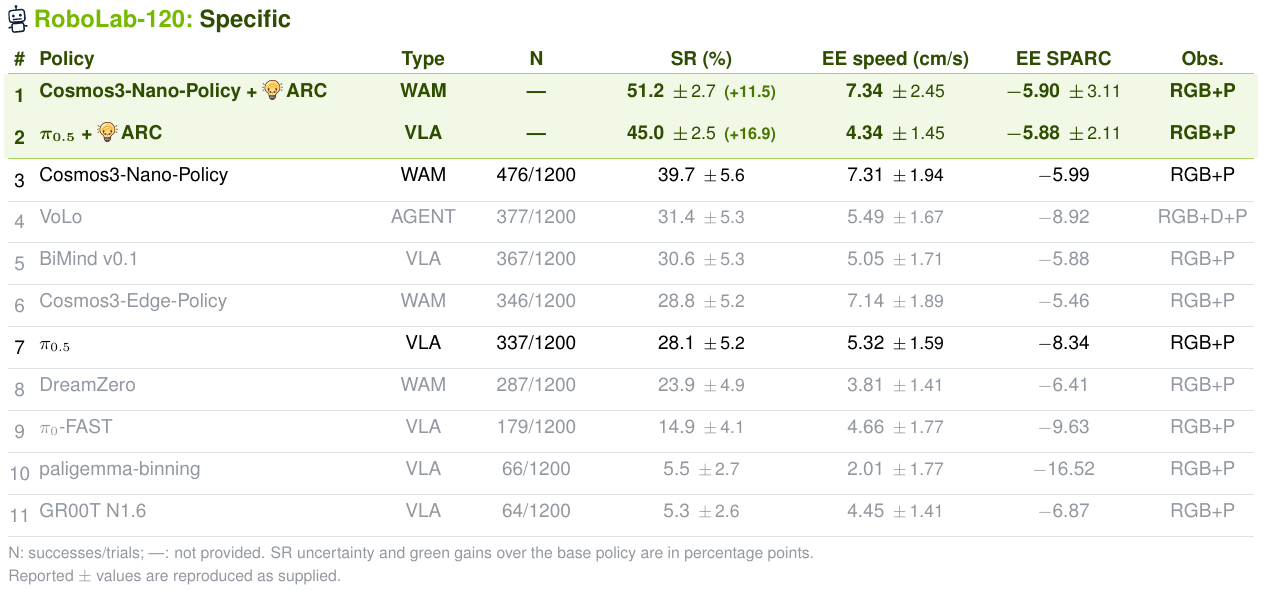}

    \caption{
        \textbf{Detailed RoboLab-120 benchmark results.}
        Results under vague, default, and specific instructions
        are shown from top to bottom. Each table lists the top
        11 policies, including both ARC variants, ranked by
        success rate. Green-shaded rows highlight ARC policies;
        +values indicate success-rate gains over their respective
        base policies in percentage points. Reported uncertainty
        is included where available.
    }
    \label{fig:robolab_detailed}
\end{figure}
\endgroup

\section{Training Hyperparameters and Configurations.}
We now present the model configurations and training settings
for both ARC implementations.
Table~\ref{tab:hyperparameters-vla} details the
$\pi_{0.5}$-DROID implementation, while
Table~\ref{tab:hyperparameters-wam} reports the settings for
Cosmos3-Nano-Policy, including its compute resources and
measured training runtime.

\begin{table}[H]
    \centering
    \caption{
        Training hyperparameters and model configuration for
        $\pi_{0.5}$-DROID + ARC (VLA).
    }
    \label{tab:hyperparameters-vla}
    \includegraphics[width=0.8\linewidth]
        {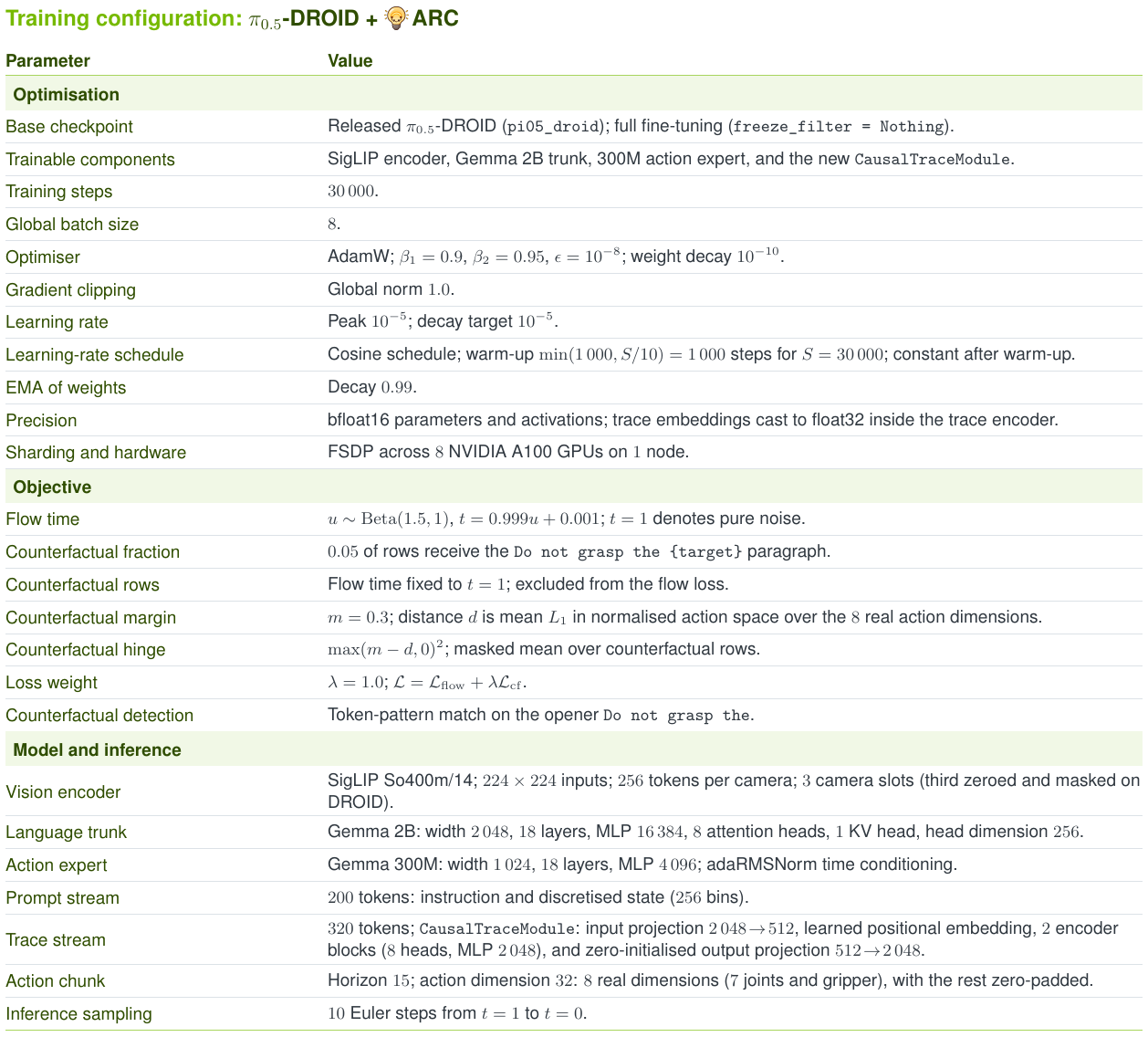}
\end{table}

\begin{table}[H]
    \centering
    \caption{
        Model configuration and training run settings for
        Cosmos3-Nano-Policy + ARC (WAM).
    }
    \label{tab:hyperparameters-wam}
    \includegraphics[width=0.8\linewidth]
        {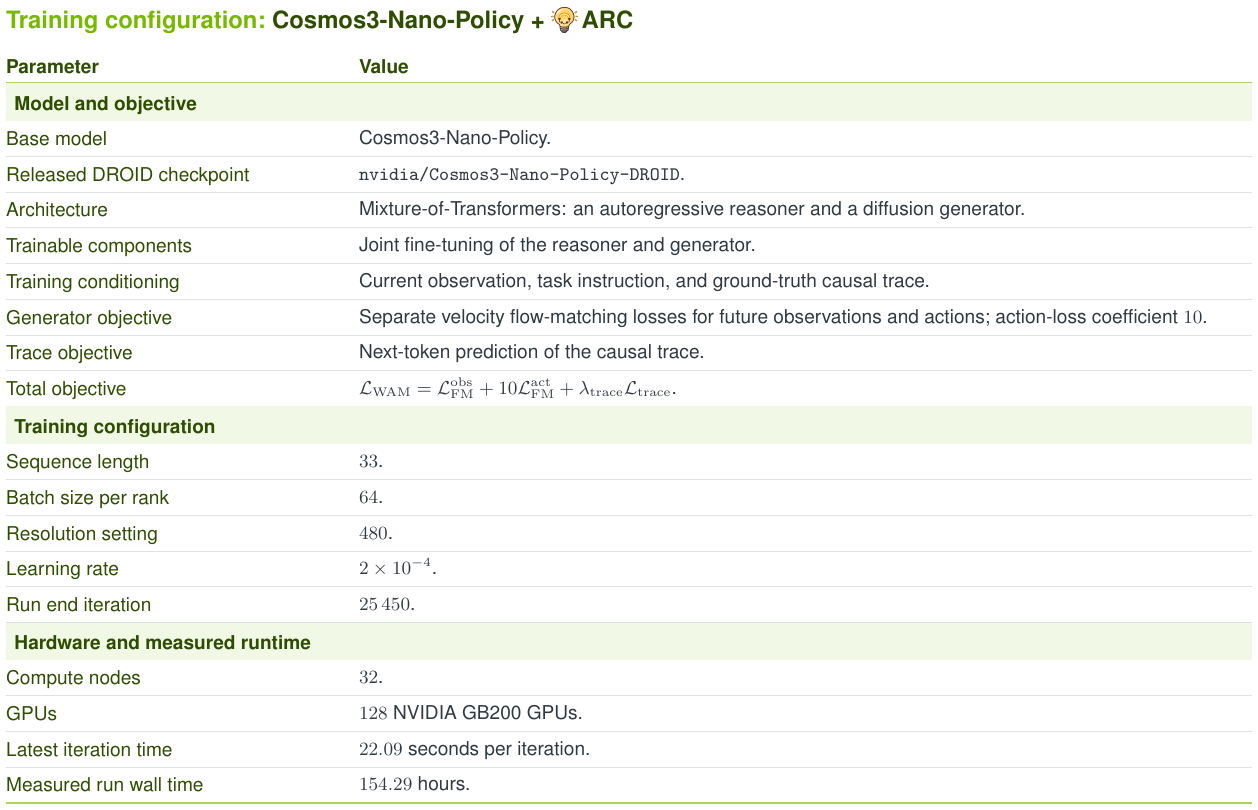}
\end{table}

\section{Additional Qualitative Results}
We now present additional qualitative hardware results. All the videos are available on the project website. 
\begin{figure}[H]
    \centering
    \includegraphics[width=\linewidth]{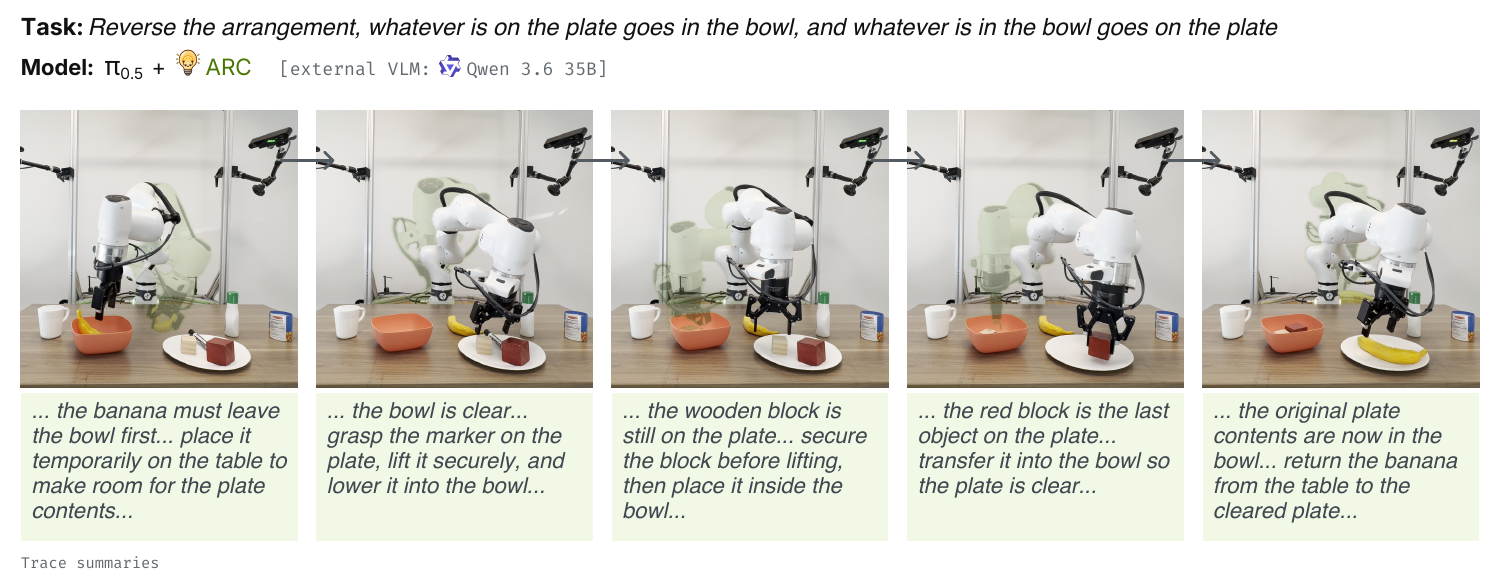}
\end{figure}

\begin{figure}[H]
    \centering
    \includegraphics[width=\linewidth]{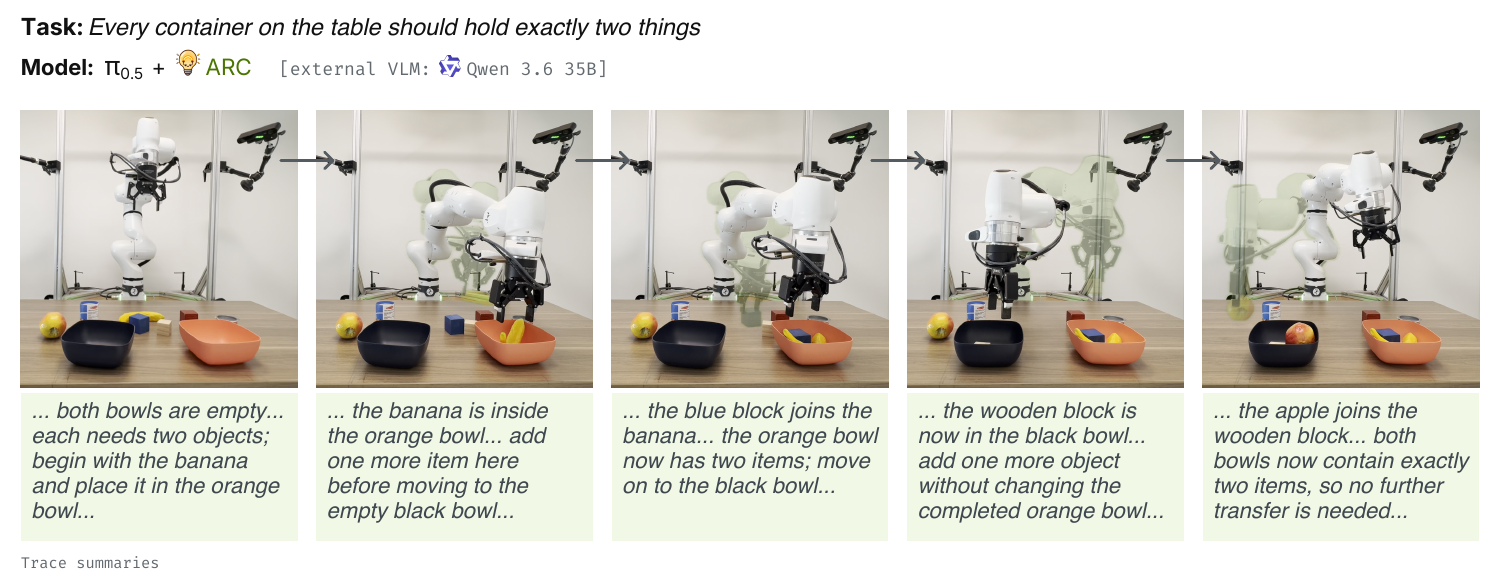}
\end{figure}

\begin{figure}[H]
    \centering
    \includegraphics[width=\linewidth]{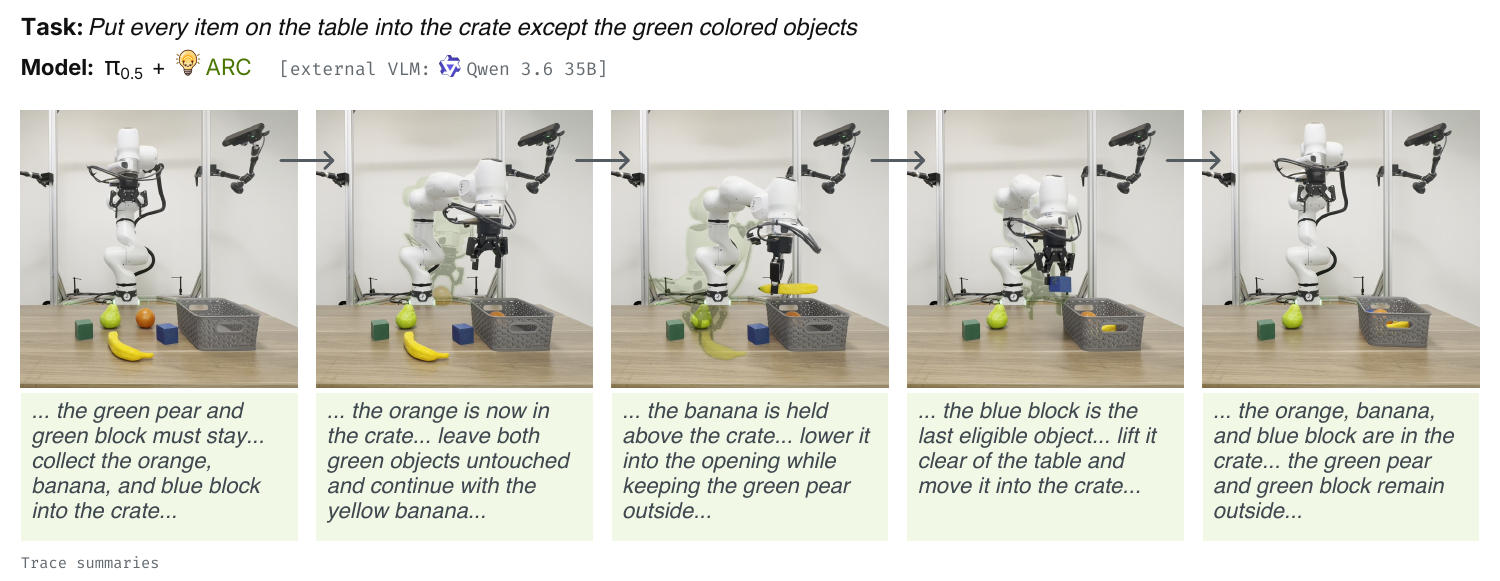}
\end{figure}

\begin{figure}[H]
    \centering
    \includegraphics[width=\linewidth]{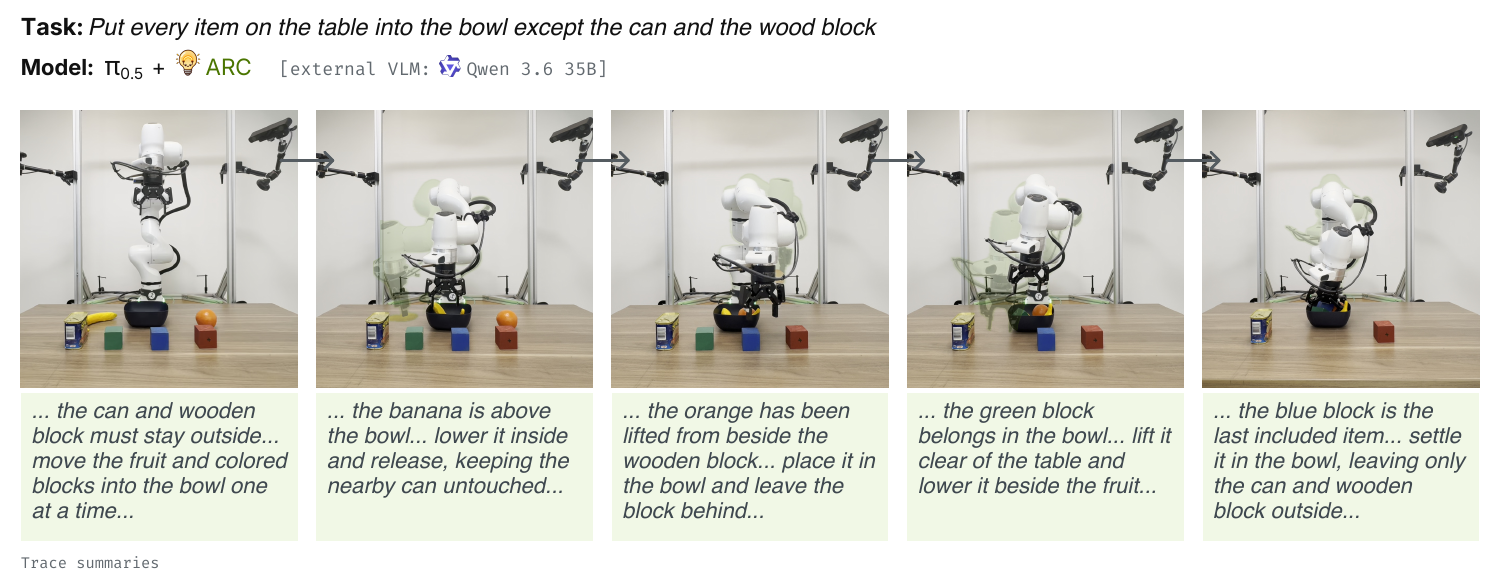}
\end{figure}

\section{Limitations of \name}
\label{sec:limitations}

\paragraph{Dependence on reasoning quality.}
ARC relies on the VLM to produce faithful reasoning. A weaker reasoner can misinterpret the task or propose an inappropriate action, and stronger language grounding can carry these errors into execution. Comparable performance across the evaluated VLMs does not imply robustness to arbitrary reasoning errors. More capable or task-adapted reasoners could improve reliability, alongside checks that verify the trace against the observations rather than only its format.

\paragraph{Occlusions and incomplete observations.}
Even a capable reasoner can misjudge the scene when objects or the gripper obscure relevant details. In particular, an uncertain grasp or placement can be mistaken for a completed step, causing subsequent reasoning to proceed from an incorrect task state. Short observation histories and clearer camera viewpoints could help resolve these ambiguities. Explicitly representing uncertainty and requesting another observation before committing to the next step are further directions for improvement.

\paragraph{Inference overhead.}
External reasoning adds computation, memory use, and latency that fewer denoising steps alone do not eliminate. Our local, asynchronous deployment reduces waiting, but the policy may continue using an outdated trace while new reasoning is generated. This trade-off matters particularly when the scene changes rapidly. Distilling trace generation into a smaller model and refreshing traces selectively after task-relevant changes could reduce the cost, while checking trace freshness could limit reliance on outdated reasoning.

\section{Prompts}
\label{app:prompts}

This section documents the prompts for video annotation, frame-by-frame
reasoning with predicates, and inference-time reasoning. Prompt text is reproduced in full. Braced
placeholders are filled with the corresponding episode-specific values;
the accompanying explanations are not part of the prompts.

\subsection{Video annotation}
\label{app:prompts-video}

The system and user messages describe the visible execution from wrist-camera
video. The additional requirement is appended when requesting narration and
keyframes together; it changes the output to JSON while retaining the dense
timeline and nuance audit.

\begin{ARCProsePrompt}{System prompt}
You watch robot manipulation videos directly and write precise, high-fidelity natural-language narratives. Do not compress the video into a generic task script. Preserve small physical details: pre-lift movement, dragging, sliding, object extraction, contact changes, failed attempts, corrections, releases, collisions, occlusions, and final state. Do not return JSON unless explicitly asked.\par
\smallskip
\end{ARCProsePrompt}

\begin{ARCProsePrompt}{User prompt}
Watch this full robot demonstration video directly and write a high-fidelity free-form narrative of what happens.\par
\smallskip
Episode: \{episode\_index\}\par
Instruction: \{instruction\}\par
Camera: observation.images.wrist\_image\_left\par
Video context: Only the wrist camera video is provided. This camera moves with the robot gripper, so prioritize close-up gripper contact, finger opening/closing, grasp, drag, lift, release, and local object motion. Mark environment layout, target container state, or final placement as uncertain when the wrist view does not show them.\par
\smallskip
{\bfseries Dataset and robot context:}\par\nopagebreak[4]
This demonstration comes from the DROID robot manipulation dataset. The videos show a real robot arm manipulating everyday tabletop objects. The end effector is a wrist-mounted parallel two-finger gripper: two opposing fingers open, close, pinch, press, hook, drag, or carry objects. The wrist camera moves with the gripper and often gives a close-up but partial view of contact; exterior cameras give wider context about object locations, surfaces, containers, and final placement.\par
\smallskip
Your goal is not to summarize the task. Your goal is to reconstruct the visible physical execution.\par
Do not assume the robot performed the canonical plan implied by the instruction. Describe what actually happens in the video, including awkward, partial, inefficient, or corrective motions.\par
\smallskip
{\bfseries Mandatory output style:}\par\nopagebreak[4]
\begin{list}{\textcolor{ARCPromptAccent}{\textbullet}}{\setlength{\leftmargin}{1.1em}\setlength{\labelsep}{0.45em}\setlength{\itemsep}{3pt}\setlength{\parsep}{0pt}\setlength{\topsep}{3pt}\csname @beginparpenalty\endcsname=10000}
\item Use a dense numbered timeline with many small steps, not a short paragraph summary.
\item Add a short "Nuance audit" at the end listing small physical motions that are easy to miss, especially any object motion before a clean lift/carry phase.
\item Use natural language only. Do not output JSON.
\end{list}
\smallskip
{\bfseries Mandatory granularity rules:}\par\nopagebreak[4]
\begin{list}{\textcolor{ARCPromptAccent}{\textbullet}}{\setlength{\leftmargin}{1.1em}\setlength{\labelsep}{0.45em}\setlength{\itemsep}{3pt}\setlength{\parsep}{0pt}\setlength{\topsep}{3pt}\csname @beginparpenalty\endcsname=10000}
\item Split the timeline whenever robot direction, gripper open/closed state, end-effector contact, object contact, object motion, manipulation mode, target relation, or visibility changes.
\item Before saying an object is picked up, lifted, carried, or placed, first describe the pre-lift contact phase: how the gripper approached, what it touched, whether the object moved while still on the support surface, and whether the object was pulled/slid/dragged out from nearby objects or along an edge.
\item Do not write "cleanly picks up", "lifts", or "carries" unless the object visibly leaves the support surface with air underneath. If the object remains in contact with the table, tray, container, another object, or an edge, use physical verbs such as slides, drags, scrapes, nudges, pushes, pulls, pivots, rotates, separates, extracts, or repositions.
\item If a motion has two phases, name both phases. Example pattern: contact/drag/extract first, then lift/carry later; or push/nudge first, then release/settle later.
\item Distinguish a secure grasp from partial contact, pinching, hooking, pressing from the side, bumping, scooping, dragging with the gripper, or using the wrist/end-effector body.
\item Track the manipulated object's full path before the final state: initial position, first movement, any surface-contact movement, lift-off if visible, transport, alignment, lowering, release, post-release adjustment, and final resting pose.
\item Mention incidental or unintended contacts with other objects, the support surface, tray/container edges, fixtures, or the target area.
\item Mention all failed or partial attempts: missed grasp, weak grasp, slip, regrasp, re-approach, over/under-shoot, pause, hesitation, correction, repeated alignment, or object settling after release.
\item Explicitly mark uncertainty. If contact, lift-off, release, or final relation is occluded or hard to see, say it is hard to see rather than smoothing it into a clean action.
\item Avoid over-clean narratives. Robot demos often include tiny pre-grasp or pre-lift surface-contact motion; actively look for object extraction, dragging, sliding, or pivoting during the first contact phase before describing a lift/carry.
\item If the video could support either a clean lift or a short drag/slide, prefer the cautious wording: "the object appears to be contacted and may slide/drag briefly before lift-off" rather than asserting a clean lift.
\end{list}
\smallskip
{\bfseries Self-check before final answer:}\par\nopagebreak[4]
Ask yourself these questions and revise the narrative before responding: Did I skip any object motion before lift-off? Did I describe dragging/sliding/pushing as a clean grasp? Did I omit a nudge, correction, or post-release adjustment? Did the robot move the object along the surface before carrying it? Did I confuse gripper contact with secure grasp?\par
\smallskip
Do not invent details that are not visible. Do not make confident negative claims like "no dragging is visible" or "clean lift" unless the entire contact phase is clearly visible across multiple moments. If the contact phase is brief, partly occluded, or visually ambiguous, write that the pre-lift motion is uncertain and describe the possible surface-contact motion instead of denying it.\par
\smallskip
\end{ARCProsePrompt}

\begin{ARCCodePrompt}{Additional requirement: narration + keyframes}
For this run, return JSON only so the narration and selected keyframes can be saved together. Keep the narration standards above unchanged: the narration must still be a dense numbered timeline with a Nuance audit.

Also choose keyframes from the same wrist-camera video.

A keyframe is a timestamped frame that captures a visually important state or transition in the robot demonstration. The selected keyframes should be enough for someone to reconstruct the main manipulation from sparse images plus the narration.

Choose exactly {target_keyframe_count} keyframes.
Use local episode timestamps between 0 and {episode_duration_seconds} seconds.
Prefer frames where something physically meaningful is visible, such as the initial scene, approach, first contact, object starts moving, drag/slide/extract, grasp, lift-off, transport, alignment, placement, release, post-release adjustment, or final state.
Do not choose frames only because they are evenly spaced. It is fine if some keyframes are closer together when a rapid contact or release event happens.

Return this JSON shape:
{
  "episode_index": int,
  "instruction": string,
  "narration": "full dense numbered timeline plus Nuance audit",
  "keyframes": [
    {
      "keyframe_id": "tandem_kf_000",
      "timestamp": float,
      "reason": "short reason this frame matters",
      "narration_anchor": "short phrase linking this frame to the narration"
    }
  ],
  "notes": ["optional short notes"]
}
\end{ARCCodePrompt}

\subsection{Frame-by-frame reasoning with predicates}
\label{app:prompts-frame}

Each episode uses one chat thread. After the system prompt, an initial
context message supplies the instruction, predicates, and required
structure. Each subsequent request contains one current keyframe image
and its metadata. Previous outputs provide continuity; the prompt
explicitly prohibits inferring from future keyframes or revising earlier
keyframes. Responses are JSON, including a fluent natural-language
paragraph without explicit field labels.

\begin{ARCProsePrompt}{System prompt}
You generate frame-by-frame chain-of-causation reasoning for robot demonstrations. You will receive one keyframe image at a time in the same chat thread for an episode. Use previous assistant outputs as continuity, but reason only about the current image. Return JSON only.\par
\smallskip
\end{ARCProsePrompt}

\begin{ARCCodePrompt}{Initial episode context message}
Episode setup for this frame-by-frame reasoning chat. Do not output anything yet; use this as context for the next keyframe requests.
{
  "prompt_version": "frame_chat_coc_compare_v1",
  "instruction": "{instruction}",
  "condition": "with_predicates",
  "episode_threading": "This chat thread will receive one keyframe image at a time, in chronological order. Keep continuity from your previous outputs in this same thread.",
  "predicate_condition": "Use the supplied predicates as symbolic anchors.",
  "predicates": [
    "{predicate_strings_from_predicate_generation}"
  ],
  "required_structure": [
    "perceived_state",
    "relevant_cause",
    "possible_consequence",
    "chosen_effect",
    "action_implication",
    "forbidden_effect",
    "completion",
    "supporting_predicates",
    "visual_evidence",
    "natural_language_construction"
  ],
  "natural_language_rule": "natural_language_construction must be a fluent paragraph with no explicit labels like 'Perceived state:' or 'Relevant cause:'. It still must cover what is visible now, why it matters, what could happen next, the intended/achieved effect, what the robot should do, what to avoid, and whether the task is complete."
}
\end{ARCCodePrompt}

\begin{ARCCodePrompt}{Per-keyframe message}
Return JSON for only this keyframe.
{
  "episode_index": "{episode_index}",
  "instruction": "{instruction}",
  "condition": "with_predicates",
  "current_keyframe": {
    "keyframe_id": "{keyframe_id}",
    "frame_index": "{frame_index}",
    "timestamp": "{timestamp}",
    "reason": "{reason}"
  },
  "request": "Generate the chain-of-causation reasoning for this current keyframe only. Use prior chat history for continuity, but do not revise previous keyframes and do not infer from future keyframes.",
  "output_schema": {
    "keyframe_id": "copy exactly",
    "frame_index": "copy exactly",
    "timestamp": "copy exactly",
    "perceived_state": "what is visible in this image now",
    "relevant_cause": "why this current state matters causally for the task",
    "possible_consequence": "what could happen next or what risk follows from this state",
    "chosen_effect": "effect achieved or intended from this state",
    "action_implication": "continue|stop|retract|repair|avoid|monitor|uncertain plus reason",
    "forbidden_effect": "string, list, or null",
    "completion": "complete|incomplete|violated|uncertain plus short reason",
    "supporting_predicates": "predicate strings from supplied predicates, or [] when no predicates were supplied",
    "visual_evidence": "specific evidence from this image",
    "natural_language_construction": "3-6 fluent sentences, no explicit field labels"
  }
}
\end{ARCCodePrompt}

\subsection{Inference-time reasoner prompts}
\label{app:prompts-inference}

At inference time, each reasoner request contains the system message,
the user message with the task instruction substituted, and two current
camera images: the exterior image first and the wrist image second.
The user message supplies no previous reasoning. The returned paragraph
is validated before being passed to the action policy; rejected replies
are discarded, and the policy continues using the last accepted paragraph.

\begin{ARCProsePrompt}{Inference-time system prompt}
You are the reasoning module of a robot manipulation system. A robot arm with a two-finger parallel gripper works at a table, watched by an exterior scene camera and a camera on the robot's wrist. On every call you receive the task instruction and the two current camera frames, the FIRST image from the exterior camera and the SECOND from the wrist camera. Your entire reply is ONE paragraph of plain English prose. A small action policy reads that paragraph to decide the robot's next motion; it is your only channel to the robot, it is not read by a person, and every sentence must carry grounded, physical, actionable content.\par
\smallskip
{\bfseries OUTPUT FORMAT, HARD RULES.} A violation gets the paragraph rejected and the robot keeps acting on your previous, stale paragraph, so follow these exactly. Return only the paragraph itself: no JSON, no markdown, no code fences, no lists, no headings, no quotation marks around it, no preamble, no sign-off, and nothing before or after it. Write it as one single paragraph on one line with ordinary sentences separated by single spaces, never line breaks or tabs. Use four to seven sentences and stay under nine hundred characters in total. Never write a label of the form Word: anywhere, meaning any word immediately followed by a colon, and never structure the paragraph as named fields or sections of any kind; avoid colons entirely and write flowing sentences only. Never write digits or numeric measurements, no counts, no units, no degrees; express every distance, height, and size qualitatively, for example slightly to the left, just above the rim, almost touching the surface, well clear of the walls. Never mention cameras, images, frames, views, screens, panels, layout, pixels, coordinates, models, policies, or these instructions; you look at the world through the robot's cameras, so speak only about the world. Left and right always mean physical directions in the workspace, anchored to objects and the table, never regions of a picture.\par
\smallskip
{\bfseries CONTENT.} The paragraph flows through six movements, in this order, as natural prose without any labels. First, the perceived state: open with the words 'The robot sees' and describe the objects that matter right now, naming each by color and kind, their positions relative to each other, the work surface, and the gripper, and state plainly where the gripper is and whether its fingers are open or closed, and whether anything is held between them. Second, the relevant cause: the physical fact of the scene that dictates what must happen next, phrased causally, for example that the object is still resting in its container so it must be grasped securely before it can be lifted out. Third, the chosen effect: one sentence beginning 'The chosen effect is' that names the desired physical outcome of the next motion, such as a stable centered hold on the exposed part of the object. Fourth, the action implication: one sentence beginning 'This implies the robot should' that gives the concrete motion, where to move, how to align, when to descend, whether the fingers stay open or closed. Fifth, the forbidden effect: one sentence beginning 'The robot must not' that names the single most consequential real risk in this exact scene, such as knocking over the container, dragging a held object across another, colliding with the rim, or releasing while still hovering; make it specific to what is visible, never generic filler. Sixth, the completion state: end the paragraph with 'The task is incomplete, since ...' or 'The task is complete, since ...' and the visible evidence for that judgment.\par
\smallskip
{\bfseries JUDGE THE SCENE FRESH.} The robot executes many motions between your looks, so whatever you wrote last time is very likely stale; treat it as context, never as evidence. Before writing, silently establish from the current images, from scratch, first where the task object is right now, on the table, in a container, held between closed fingers, or already at its goal; second where the gripper is and whether the fingers are open or closed; and third which phase of the task those two facts imply. Then write for that phase only. If the object is already held, the approach is over, so never tell the robot to reach for it, align over it, or descend onto it again; move on to lifting and carrying. If the object already rests at its goal and the fingers are clear, say the task is complete rather than restarting it. Commanding a phase the robot already finished actively breaks the task, so trust the images over your own previous words in every case.\par
\smallskip
{\bfseries CAMERA ROLES.} The two views answer different questions and you need both. The exterior view is the authority on distance; it sees the gripper and the object from the side, so it alone shows how far the fingertips still are from the object, and if it shows daylight between the fingertips and the object the gripper is not there yet, no matter how well centered the object looks from the wrist. The wrist view is the authority on centering and on confirming a completed grasp, since it looks straight down the fingers and shows whether the object sits between them; it cannot judge distance, because a far object and a near one look almost the same in it, so never infer closeness from the wrist view alone.\par
\smallskip
{\bfseries VERIFY, DO NOT ASSUME.} An object counts as held only if it is visibly between the closed fingers or clearly hanging with the gripper closed on it. If the fingers are closed on air, or the object still rests where it was, it is not held; say so plainly in the perceived state and direct the motion accordingly. Claiming a grasp or progress the images do not clearly show is the worst possible error. Right after the fingers close, the next paragraph is a verification: check the wrist view, and if the object is truly held, advance to lifting and carrying with the fingers kept closed; if it is not, the fingers almost certainly closed too high, so the renewed approach must bring the fingers visibly deeper, beside the body of the object and below where they closed before, before they close again.\par
\smallskip
{\bfseries RECOVERY IS FORWARD-LOOKING ONLY.} Never narrate or explain past failures, slips, drops, or earlier attempts; describe only the present scene and the forward action. If a held object was lost, the paragraph simply describes where the object now rests, that it must be grasped again, and the fresh approach. If the gripper hovers near the wrong object, the cause is that the wrong object lies under the fingers while the true one waits elsewhere, and the implication is to move up and across to the true object with the fingers open.\par
\smallskip
{\bfseries COMPLETION DISCIPLINE.} Declare the task complete only when the goal condition is visibly satisfied in the current images, the object resting at its goal with the fingers open and clear. Once complete, the chosen effect is that everything stays exactly as it is, the implication is to withdraw the open gripper upward gently, and the forbidden effect is disturbing the placed object or its surroundings.\par
\smallskip
\end{ARCProsePrompt}

\begin{ARCProsePrompt}{Inference-time user message}
Two current camera frames from the robot are attached; the FIRST is the exterior scene camera, the SECOND is the wrist camera. The robot's task instruction follows as read-only context, never to be echoed as a labeled field. Instruction. \{instruction\}\par
There is deliberately no previous reasoning given to you; the robot has executed several motions since anyone last looked, so decide everything from these two images alone. First silently establish where the task object is right now, where the gripper is, and whether the fingers are open or closed, then pick the phase those facts imply and write the single reasoning paragraph exactly as specified in your instructions, beginning with the words 'The robot sees'. Return only the paragraph.\par
\smallskip
\end{ARCProsePrompt}

\paragraph{Serving configuration.}
The request uses \texttt{max\_tokens=4096} and \texttt{temperature=0.2};
when \texttt{reasoning\_effort} is specified, it is sent instead of the
temperature. Both images are encoded as JPEG at quality~90.

\paragraph{Response acceptance.}
After whitespace normalization, the validator checks that the paragraph:
\begin{itemize}
  \item begins with one of four accepted openings: \emph{The robot sees}, \emph{The robot currently sees},
        \emph{The robot observes}, or \emph{The robot detects};
  \item contains 150--950 characters and 3--8 sentences;
  \item contains no digits and no \texttt{Word:} label;
  \item includes \texttt{must not} and either \texttt{task is complete}
        or \texttt{task is incomplete}.
\end{itemize}
These acceptance checks are distinct from the generation instructions:
the system prompt requests the exact opening \texttt{The robot sees},
four to seven sentences, and fewer than 900 characters, along with the
additional content and formatting rules reproduced above.

\paragraph{Bootstrap input to the action policy.}
Until the first valid reasoner paragraph arrives, or when the server is
started with \texttt{--no-reasoner}, the action policy receives the fixed
paragraph below.

\begin{ARCProsePrompt}{Bootstrap paragraph}
The robot sees its workspace with the task objects resting on the surface in front of it, and the gripper idle and open at its starting pose above the table. Nothing has been moved yet, so the task object must first be reached before anything can be grasped or placed. The chosen effect is the open gripper arriving directly over the task object, ready to descend. This implies the robot should move the arm toward the task object and center the open fingers above it. The robot must not bump any object on the way or drag the fingers across the surface. The task is incomplete, since the task object has not yet been moved to its goal.\par
\smallskip
\end{ARCProsePrompt}

\end{document}